\pdfoutput=1

\documentclass[11pt]{article}

\usepackage{acl}

\usepackage{times}
\usepackage{latexsym}

\usepackage[T1]{fontenc}

\usepackage[utf8]{inputenc}

\usepackage{microtype}

\usepackage{inconsolata}

\usepackage{amsmath}

\usepackage{cleveref}

\crefname{figure}{Fig.}{Figs.}

\crefname{appendix}{App.}{App.}

\crefname{section}{Sec.}{Secs.}
\usepackage{caption}

\usepackage{subcaption}
\usepackage{tabularx}
\usepackage{booktabs}
\usepackage{multirow}
\usepackage[table]{colortbl}
\usepackage{array}
\usepackage{adjustbox}
\usepackage{hyperref}

\usepackage{afterpage}
\usepackage{pgfplots}
\pgfplotsset{compat=1.18} 
\usepackage{soul}
\usepackage{enumitem}  
\definecolor{purple}{HTML}{332288}
\definecolor{green}{HTML}{117733}
\definecolor{teal}{HTML}{44AA99}
\definecolor{cb-lilac}      {RGB}{182, 109, 255}
\definecolor{lightblue}{HTML}{56B4E9}
\definecolor{beige}{HTML}{DDCC77}
\definecolor{salmon}{HTML}{CC6677}
\definecolor{pink}{HTML}{AA4499}
\definecolor{maroon}{HTML}{882255}
\definecolor{rpurple}{HTML}{CC79A7}
\definecolor{orange}{HTML}{E69F00} 
\definecolor{cb-burgundy}   {RGB}{146,   0,   0}
\definecolor{cb-clay}       {RGB}{219, 209,   0}
\definecolor{cb-brown}      {RGB}{146,  73,   0}
\definecolor{cb-green-sea}  {RGB}{  0, 146, 146}
\definecolor{cb-purple}     {RGB}{ 73,   0, 146}
\definecolor{cb-salmon-pink}{RGB}{255, 182, 119}
\definecolor{cb-dblue}{rgb}{0.37, 0.62, 0.63}

\newcommand{\markerrword}[1]{{\color{salmon} \ul{\textbf{#1\textsuperscript{W}}}}}
\newcommand{\markerrname}[1]{{\color{cb-dblue} \ul{\textbf{#1\textsuperscript{N}}}}}
\newcommand{\markerrdate}[1]{{\color{cb-purple} \ul{\textbf{#1\textsuperscript{D}}}}}
\newcommand{\markerrnumber}[1]{{\color{cb-burgundy} \ul{\textbf{#1\textsuperscript{U}}}}}
\newcommand{\markerrother}[1]{{\color{maroon} \ul{\textbf{#1\textsuperscript{O}}}}}
\newcommand{\markerraddition}[1]{{\color{lightblue} \ul{\textbf{#1\textsuperscript{A}}}}}

\newcommand{\markerrcontext}[1]{{\color{cb-lilac} \ul{\textbf{#1\textsuperscript{C}}}}}
\newcommand{\markerrnoneng}[1]{{\color{cb-clay} \ul{\textbf{#1\textsuperscript{NE}}}}}
\newcommand{\markerrnotcheck}[1]{{\color{green} \ul{\textbf{#1\textsuperscript{NC}}}}}

\usepackage{comment}
\usepackage{booktabs}
\usepackage{todonotes}
\usepackage{algorithm}

\usepackage{algpseudocode}
\definecolor{red}{HTML}{DD0000}

\newcommand{\lineacross}{\rule{\linewidth}{1pt}}

\usepackage{array}
\newcolumntype{P}[1]{>{\centering\arraybackslash}p{#1}}
\usepackage[page]{appendix}
\usepackage{subcaption}

\newcommand{\customsize}{\fontsize{8}{11}\selectfont}
\title{Improving Factual Accuracy of Neural Table-to-Text Output by Addressing Input Problems in ToTTo}

\author{Barkavi Sundararajan \and Somayajulu Sripada \and Ehud Reiter\\
  Department of Computing Science, University of Aberdeen \\
  \texttt{\{b.sundararajan.21, yaji.sripada, e.reiter\}}{@abdn.ac.uk} \\}

\begin{document}
\maketitle
\begin{abstract}
 
Neural Table-to-Text models tend to hallucinate, producing texts that contain factual errors.  We investigate whether such errors in the output can be traced back to problems with the input.  We manually annotated 1,837 texts generated by multiple models in the \emph{politics} domain of the ToTTo dataset.  We identify the input problems that are responsible for many output errors and show that fixing these inputs reduces factual errors by between 52\% and 76\% (depending on the model).  In addition, we observe that models struggle in processing tabular inputs that are structured in a non-standard way, particularly when the input lacks distinct row and column values or when the column headers are not correctly mapped to corresponding values.

\end{abstract}

\section{Introduction}

Table-to-Text generation refers to the task of generating natural language descriptions from tabular data \citep{parikh-etal-2020-totto, DBLP:journals/corr/abs-2004-10404, chen-etal-2020-logic2text} and is widely used in several application domains such as medical diagnosis \citep{Pauws2018MakingEU}, financial \citep{10.1145/3543873.3587598} and weather reporting \citep{sripada2002sumtime, 7983470, upadhyay-massie-2022-content} and sports summaries \citep{thomson-etal-2020-sportsett}. Neural language models are known to generate fluent texts \citep{ji2023survey} but may generate outputs that are factually incorrect or unrelated to the provided input data. Such undesirable generation is called `hallucination' \citep{wang-sennrich-2020-exposure, raunak-etal-2021-curious, ji2023survey}.  

Previous studies on Table-to-Text tasks adopt traditional seq2seq methods to generate table descriptions \citep{wiseman-etal-2017-challenges, puduppully-et-al-2019-planning, Rebuffel2019AHM}. Recently, Transformer based models \citep{Devlin2019BERTPO, 2020t5, OpenAI2023GPT4TR} have shown remarkable progress in language generation from textual input \citep{badaro-etal-2023-transformers}, however tabular data still needs more improvement to control hallucinations \citep{Rebuffel2021ControllingHA}. Neural models struggle with tabular data, especially when the inputs do not have distinct cell values from rows and columns mapped along with their respective headers. These input problems lead the model to generate more factual errors \citep{kasner2024beyond}. 

Using the ToTTo tabular dataset \citep{parikh-etal-2020-totto}, we identify and address input problems that are responsible for factual errors. Some common tabular input problems in the ToTTo are \textbf{i.} `non-atomic' cell values, where a column contains multiple values such as leader name, party name and \% of votes in one cell rather than a single indivisible value, \textbf{ii.} missing important cell values in the input (see \Cref{tab:intro_problem}) and \textbf{iii.} nested column headers and row headers in the Wikipedia tables\footnote{\url{https://en.wikipedia.org/wiki/Wikipedia:Manual_of_Style/Accessibility/Data_tables_tutorial\#Column_headers:_bad_example}} that lead to incorrect mapping of the cell values.

\begin{table*}[htbp]
    \lineacross{}
    \setlength\extrarowheight{1pt}
    \footnotesize

    \definecolor{yellow}{rgb}{1,1,0.5}
    \definecolor{cby}{rgb}{0.91, 0.84, 0.42}
    \definecolor{cbora}{RGB}{255,217,191}
    \definecolor{red}{HTML}{DD0000}

     \begin{subtable}[ht]{0.50\textwidth}
        \vspace{-12px}
        \subcaption{\textbf{Original cells highlighted in Yellow as Input}}
        \label{tab:a}
        \vspace{2px}
        \begin{tabular}{p{1.5cm}|p{2.5cm}|p{1cm}|p{0.8cm}}
            \multicolumn{4}{p{7cm}}{\textbf{1996 United States House of Representatives election}}\\
            \hline
            \textbf{Party} & \textbf{Candidate} & \textbf{Votes }& \textbf{\%} \\
            \hline
            Democratic & \cellcolor{yellow}Eleanor Holmes Norton (inc.) & 134,996 & 90.00 \\
            \hline
            \cellcolor{yellow}Republican & \cellcolor{yellow}Sprague Simonds & 11,306 & \cellcolor{yellow}7.54 \\ [1ex]
            \hline 
        \end{tabular}
    
        \vspace{4px}
        \begin{tabular}{p{7.5cm}}
            {\textbf{\textsc{Llama 2-13B} Output for tabular input a:}} \\
            Eleanor Holmes Norton (inc.) won with \markerrnumber{7.54\%} of the vote. Sprague Simonds was the Republican candidate and received \markerrnumber{22.38\%} of the vote.
        \end{tabular}

    \end{subtable}
    \begin{subtable}[ht]{0.52\textwidth}
    \vspace{-12px}
    \subcaption{\textbf{Corrected Tabular data by including relevant cells}}
    \label{tab:b}
           \vspace{2px}
           
           \begin{tabular}{p{1.5cm}|p{2.5cm}|p{1cm}|p{0.8cm}}
            \multicolumn{4}{p{7cm}}{\textbf{1996 United States House of Representatives election}}\\
            \hline
            \textbf{Party} & \textbf{Candidate} & \textbf{Votes }& \textbf{\%} \\
            \hline
            \cellcolor{yellow}Democratic & \cellcolor{yellow}Eleanor Holmes Norton (inc.) & 134,996 & \cellcolor{yellow}90.00 \\
            \hline
            \cellcolor{yellow}Republican & \cellcolor{yellow}Sprague Simonds & 11,306 & \cellcolor{yellow}7.54 \\ [1ex]
            \hline
            \end{tabular}
        
            \vspace{4px}
            \begin{tabular}{p{7.5cm}}
                \textbf{\textsc{Llama 2-13B} Output for tabular input b:} \\
                Democratic Party candidate Eleanor Holmes Norton won with 90\% of the vote. Republican Party candidate Sprague Simonds received 7.54\%. \\ [1ex]
            \end{tabular}
    \end{subtable}
    \vspace{-4px}
    \lineacross{}
    \caption{\textbf{ToTTo example: Highlighted cells in yellow are passed as input to the model.} Passing the appropriate cells (i.e., \% votes and party name) as input, as shown in \Cref{tab:b} fixes the factual errors. Compare the \Cref{tab:b} output (with no errors) to the \Cref{tab:a} output (with NUMBER errors denoted by a superscript U).} 
    \label{tab:intro_problem}
    
    \vspace{4px}
    \lineacross{}
    \begin{tabular}{p{15.6cm}}
    <page\_title> 1996 United States House of Representatives election </page\_title> <table> <cell> Eleanor Holmes Norton (inc.) <col\_header> Candidate </col\_header> </cell> <cell> Republican <col\_header> Party </col\_header> </cell> <cell> Sprague Simonds <col\_header> Candidate </col\_header> </cell> <cell> 7.54 <col\_header> \% </col\_header> </cell> </table> \\
    \end{tabular}
    
    \vspace{2px}
    \lineacross{}
    \caption{\textbf{Linearized Input from (a) and (b) is passed as input to \textsc{Llama 2-13B} model.} For example, we present the Linearized Input for \Cref{tab:a} here. The corresponding input for \Cref{tab:b} is not shown, but it will include all related cells highlighted in yellow in (b).}
    \label{tab:input_text}
    
\end{table*}

\Cref{tab:intro_problem} presents a sample from ToTTo. Only the highlighted cells from \Cref{tab:a} are passed to the model (as shown in \Cref{tab:input_text}). Passing Norton's \% of votes and her party name (compare \Cref{tab:b} to \Cref{tab:a}) eliminates the hallucinated \% of votes (see output for \Cref{tab:b}); this correction is based on the input problem described in  \Cref{subsec: fixes}.

In this paper, we score the quality of output texts by manually annotating output errors instead of using automatic evaluation metric scores such as BLEU \citep{papineni-etal-2002-bleu}, ROUGE \citep{lin-2004-rouge}, PARENT \citep{dhingra-etal-2019-handling} and BLEURT \citep{sellam-etal-2020-bleurt}. We conducted a pilot study, where we fine-tuned T5-base and T5-large models \citep{2020t5}, analysing 1,677 \emph{politics} domain texts from the ToTTo dataset through manual error analysis adopted from \citet{thomson-reiter-2020-gold}. These manual error annotations allowed us to identify patterns of errors in the generated text which were then traced back to input problems.  

Our approach is summarised as follows: 

\begin{enumerate}[label=\Roman*.]
  \item We systematically correct the tabular inputs for the \emph{politics} domain in ToTTo to adhere to a standard form to ensure the generation of factual texts from neural models. The correction procedure is elaborated in \Cref{subsec: fixes} and is supplemented by pseudocode in \Cref{appen: pseudo_code}.
  \begin{enumerate}
    \item We apply this correction to a larger subset of 210 samples, resulting in a 62\% decrease in factual errors for T5-base and a 57\% decrease for T5-large in the generated text (\Cref{subsec: t5_results}). 
    \item We conduct experiments on \textsc{Llama 2-7B} and \textsc{Llama 2-13B} models \citep{touvron2023llama} with 40 challenging samples selected from the previous 210 samples. Tailoring zero-shot prompts for specific input and error annotation on 160 texts showed that correcting input reduces factual errors by 52\% in \textsc{Llama 2-7B} and 76\% in \textsc{Llama 2-13B} (\Cref{subsec: Llama 2_results}). 
    \end{enumerate}
    \item The manual error annotation methodology adopted from \citet{thomson-reiter-2020-gold} is detailed in \Cref{appendix b:anno_procedure}; this builds on the work of \citet{sundararajan-etal-2022-error} for ToTTo \emph{politics} domain outputs\footnote{Error annotation guidelines and sample annotations from our human evaluation is available at \url{https://github.com/BarkaviSJ/totto_politics_human_annotations}}. The inter-annotation agreement on the error annotation was good, with a Fleiss' Kappa of 0.622 (\Cref{sec: inter-anno}). 
\end{enumerate}

\subsection{Table to Text dataset, ToTTo} 
ToTTo is an open-domain English language dataset \citep{parikh-etal-2020-totto}, where the input \( X \) is taken from Wikipedia table \( T \), which includes the table's metadata (such as title) and a set of highlighted cells, along with their headers. This structured information is flattened to create a linearized text representation of the table, as mentioned in \Cref{tab:input_text}. This crowdsourced dataset is paired with a natural language description, denoted as output \( Y \), comprising a sequence of tokens \( y_1, y_2, \ldots, y_n \), which provides a coherent summary of the content present in the input table \( X \). 

These input-output pairs from the ToTTo dataset can be used for fine-tuning or prompting the neural language models. As shown in \Cref{tab:a}, the Input X is often observed to be problematic and fixing these problems is the main focus of this paper.

\section{Related Work}
\label{sec: related_work}

\textbf{Prior work on ToTTo:} \citet{wang-etal-2022-robust} proposed a framework, LATTICE, that preserves the structural information of the cell values (tokens) within the same rows or columns, and by removing the attention mechanism flow for other unrelated tokens. \citet{hu-etal-2023-improving} incorporates content planning and generation from \citet{su-etal-2021-plan-generate} and synthetically added noisy cells in their fine-tuning regime. While these approaches are model agnostic and improved automatic metric scores such as BLEU \citep{papineni-etal-2002-bleu}, PARENT \citep{dhingra-etal-2019-handling} and BLEURT \citep{sellam-etal-2020-bleurt} by few points in the leaderboard \footnote{\url{https://github.com/google-research-datasets/ToTTo}}, the fundamental problem with the tabular input remains. 

\citet{chen-etal-2022-towards-table} acknowledged that the target cells in the ToTTo tabular input are not always highlighted in their work titled `Table Structure Understanding and Text Deliberation Approach'. They used a template to extract all facts from the raw table for 1.2K training samples (only for the inputs with rows and columns fewer than 8) and employed hierarchical multi-head attention to capture the structural information in their fine-tuning process. Though this approach promises to retain the facts from raw tables, it only addresses simpler tables with fewer than 8 rows and columns, still has limitations for longer tables, complex tabular structures and non-atomic tabular cells.

Our focus on correcting inputs aiming to achieve factually correct outputs aligns with the work of \citet{dusek-etal-2019-semantic} on the E2E dataset \citep{novikova-etal-2017-e2e}. Their study also demonstrated that improving inputs in an NLG dataset helps in improving model outputs.

\textbf{Error Analysis:} While the automatic metric scores such as BLEU, PARENT and BLEURT help evaluate the model's performance at a high level, relying solely on these metrics will not address specific weaknesses i.e., lower metric scores do not provide insights into specific error types in the output. We follow guidelines from \citet{van-miltenburg-etal-2021-underreporting} to perform error analysis in NLG systems and investigate errors in the output at a more granular level by adopting the manual error annotation approach from \citet{thomson-reiter-2020-gold, sundararajan-etal-2022-error, THOMSON2023101482}.

\citet{maynez-etal-2020-faithfulness} also emphasized automatic metrics are not sufficient to study the hallucination problem and provided a detailed study on intrinsic and extrinsic hallucination in abstractive summarization. We studied hallucination in our evaluation scheme by annotating different categories of errors in the output tokens (single token or group of tokens). Mapping our adopted methodology to \citet{maynez-etal-2020-faithfulness}'s work, intrinsic is the main error category occurring when generated outputs are not faithful to the given input. It includes \markerrword{WORD}, \markerrname{NAME}, \markerrdate{DATE\_DIMENSION}, \markerrnumber{NUMBER}, \markerrcontext{CONTEXT} and \markerrother{OTHER}) from our error categories. Extrinsic refers to our \markerraddition{ADDITION} category (see \Cref{appen: categories}). 

\textbf{LLM prompts:} Empirical evaluation of prompting strategies on the three large language models (LLMs) by \citet{sivarajkumar2023empirical} in clinical NLP found that tailoring task-specific prompt is crucial for achieving accuracy. In a study on Text-to-SQL, \citet{chang2023prompt} investigated zero-shot prompting strategies, highlighting the significance of table representation. Their findings indicated that normalized database prompts outperformed unnormalized ones; this motivates our initial step of correcting tabular inputs to a standard form. In our work, we leveraged a recent LLM, \textsc{Llama 2} \citep{touvron2023llama} and tailored our zero-shot prompt \citep{NEURIPS2022_8bb0d291} specific to the content of each table.

\section{Pilot Study}
\label{sec: first_experiment} 

\subsection{Methodology}
\label{sec: baseline}

We only look at the \emph{politics} domain on the ToTTo validation set. We build upon the work of \citet{sundararajan-etal-2022-error} to identify the causes of output errors in T5 models. In this paper, we go beyond error annotations to fix these errors in our main study, both in T5 and \textsc{Llama 2} models (detailed in \Cref{sec: main_study}). 

The error categories mentioned in \citet{sundararajan-etal-2022-error} are:  \markerrword{WORD}, \markerrname{NAME}, \markerrdate{DATE\_DIMENSION}, \markerrnumber{NUMBER}, \markerrother{OTHER}, \markerrcontext{CONTEXT}, \markerrnotcheck{NOT\_CHECKABLE} and \markerrnoneng{NON\_ENGLISH}. In this work, we excluded \markerrnotcheck{NOT\_CHECKABLE} and introduced a new error category, \markerraddition{ADDITION}, which is used when the generated text has added words or phrases that diverge from the input. Definitions for all error categories are provided in \Cref{appen: categories}.

\subsection{Insights from our Pilot Study}
\label{sec: insights}

For the pilot study, we fine-tuned both the T5-base (T5-b) and the T5-large (T5-l) models on the ToTTo dataset by following the baseline approach \citep{kale-rastogi-2020-text}. The hyperparameters and fine-tuning details for these two models are shown in \Cref{appendix a: model_spec}. 

\begin{table}[htbp]
    \raggedright
    \setlength{\arrayrulewidth}{0.5pt}
    \setlength\extrarowheight{2pt}
    \scriptsize

        
        \begin{tabular}{P{2.5cm} | P{0.8cm} P{0.5cm} | P{0.8cm} P{0.5cm} } 
        \toprule
        \multirow{2}{*}{\textbf{Category}} &
        \multicolumn{2}{c}{\textbf{T5-base (T5-b)}} &
        \multicolumn{2}{c}{\textbf{T5-large (T5-l)}} \\ \cline{2-5}
        \vspace{2px}
         &
         \textbf{Count} & \textbf{\%} & 
         \textbf{Count} & \textbf{\%} \\ [0.5ex] 
         \midrule
        \textbf{No Error} & 358 & 47 & 450 & 60 \\
         \textbf{Omissions} & 272 & 36 & 218 & 29 \\
         \textbf{Errors}  & 124 & 17 & 86 & 11 \\ [1ex]
         \midrule
         \textbf{Total Count} & 
         \multicolumn{2}{c|}{\textbf{754}} &
         \multicolumn{2}{c}{\textbf{754}} \\ [1ex]
         \bottomrule
        \end{tabular}
    \caption{\textbf{Pilot Study analysis: T5-b and T5-l Models in ToTTo Politics Domain.} `Errors' category is the main focus of our work; `omissions' are excluded.} 
    \label{tab:sentence_politics}
\end{table}

Our analysis (presented in \Cref{tab:sentence_politics}) shows that:

\textbf{No Error}: 47\% of the samples from T5-b and 60\% of the samples from T5-l are error-free.

\textbf{Omissions}: Omissions occur when the generated text fails to mention some information from the input \citep{gonzalez-corbelle-etal-2022-dealing} without making any factual errors. If the output has errors and omissions, we classify it as an error. The T5-b had 36\% omissions and T5-l had 29\%. 

\textbf{Errors}: Our analysis revealed that T5-b made factual errors in 17\% and T5-l made factual errors in 11\% of the total samples.

\textbf{Hypothesis:} Based on the insights from this analysis, we hypothesize that when tabular input data is structured in non-standard ways, models struggle to interpret these ambiguous inputs leading to generate factually inaccurate output. We test this hypothesis by addressing input problems related to non-standard tabular structures.

\section{Input Problems}
\label{sec: input_problems}

Due to the practical challenges involved in improving the tabular input for the entire dataset, which includes unique headers and tabular structures for each input, our focus is on analyzing a subset of samples containing errors. We examined 124 error samples from the T5-b and 86 error samples from T5-l models within the ToTTo \emph{politics} domain, as identified in \Cref{tab:sentence_politics}. We aim to segregate the errors originating from non-standard or illogical nested table structures. We categorize these input problems into two broad categories, as briefly elaborated upon in \cref{subsec:gen} and \cref{subsec:totto}.

\subsection{Generic Input Problems}
\label{subsec:gen}
\textbf{Non-atomic tabular cell values:} When a table cell contains multiple atomic values (see \Cref{tab:1nf_problem}). Examples of such non-atomic forms include multiple leaders' names, votes, term dates, or election years all in a single cell. We further categorize these problems into `single record lacking atomicity' and `multiple records lacking atomicity' (shown in \Cref{fig:input_problems} in \Cref{appen: eg_input_problems}) to demonstrate how the models struggle when records of multiple leaders lack atomicity.

\textbf{Complex table type:} When a table contains election results in sentence form, models struggle to interpret and generate meaningful texts because the sentence form data lacks the needed context (see \Cref{tab:complex_table_input_prob} in \Cref{appen: eg_input_problems}).

\textbf{Insufficient input:} In some cases, the necessary cells are not highlighted in the tabular input, resulting in incorrect outputs. Our analyses in \Cref{tab:intro_problem} and \Cref{tab:totto_insuf_input_prob} demonstrate that outputs become factually correct when relevant cells are included. 

\textbf{Longer table input:} Models often struggle to generate accurate texts for lengthy table inputs, especially when the data is not in a standard form and lacks clear cell relationships. 

\subsection{Politics Domain-Specific Input Problems}
\label{subsec:totto}
\textbf{Politics specific headers:} In ToTTo, the use of symbols, for example, `+' or `-' instead of having clear semantic terms like \emph{`swing percentage' or `\% change compared to previous election'} as column headers caused 5\% of errors. This lack of semantic guidance in the input made it difficult for models to accurately generate the correct output text (see \Cref{tab:swing} in \Cref{appen: eg_input_problems}).

\textbf{List of leader names in the input:} In the \emph{politics} domain, we observed a specific issue when input data contains a list of leader names. Models tend to favour the leader whose name appears first in the list (for example, either from the table title or as the first leader in highlighted cells), even if they have lost the election (see \Cref{tab:totto_specific_input_prob_order_one}, \Cref{tab:totto_specific_input_prob_order_two}, \Cref{tab:totto_specific_input_prob_order_three} in \Cref{appen: eg_input_problems}). This becomes even more challenging when leader names are associated either with missing values (opponent leader's name or vote count) in other columns in the input or when the tabular cell values are non-atomic. 

The manual fixes we applied to each of these input problems with examples are detailed in \Cref{subsec: fixes}.

\section{Main Study}
\label{sec: main_study}
\begin{table*}[htb]
    \lineacross{}
    \setlength\extrarowheight{1pt}
    \customsize

    \definecolor{yellow}{rgb}{1,1,0.5}
    \definecolor{cbora}{rgb}{1,1,0.5}
    \definecolor{red}{HTML}{DD0000}
    \vspace{-36px}
    
        \begin{subtable}[ht]{0.49\textwidth}
        \subcaption{\textbf{Input Table with Original Cells highlighted in Yellow}} 
        
        \vspace{1px}
        \begin{tabular}{l|p{5.8cm}}
            \multicolumn{2}{l}{\textbf{Page Title:} 2014 United States Senate elections} \\
            \multicolumn{2}{l}{\textbf{Section Title:} Elections leading to the next Congress } \\
            \hline
            \textbf{State} & \textbf{Candidates}\\
            \hline
            \cellcolor{yellow}Alaska & \cellcolor{yellow}Dan Sullivan (Republican) 48.0\% Mark Begich (Democratic) 45.8\%  Mark Fish (Libertarian) 3.7\% Ted Gianoutsos (Independent) 2.0\% \\
            \hline
            \cellcolor{yellow}Virginia & \cellcolor{yellow}Mark Warner (Democratic) 49.1\% Ed Gillespie (Republican) 48.3\% Robert Sarvis (Libertarian) 2.4\% \\ [1ex]
            \hline
        \end{tabular}
         \label{tab:1nf_orig}
        \subcaption{\textbf{Input Table with Corrected Cells highlighted in Yellow}} 

        \begin{tabular}{l|P{2cm}|P{1.8cm}|P{1.2cm}}
            \hline
            \textbf{State} & \textbf{Candidates} & \textbf{Party} & \textbf{\% votes} \\
            \hline
            \cellcolor{yellow}Alaska & \cellcolor{cbora}Dan Sullivan & \cellcolor{cbora}Republican & \cellcolor{cbora}48.0\% \\
            \hline
            \cellcolor{cbora}Alaska & \cellcolor{cbora}Mark Begich & \cellcolor{cbora}Democratic & \cellcolor{cbora}45.8\% \\
            \hline
            \cellcolor{yellow}Virginia & \cellcolor{cbora}Mark Warner & \cellcolor{cbora}Democratic & \cellcolor{cbora}49.1\% \\
            \hline
            \cellcolor{cbora}Virginia & \cellcolor{cbora}Ed Gillespie & \cellcolor{cbora}Republican & \cellcolor{cbora}48.3\% \\ [1ex]
            \hline
        \end{tabular}
       
       \label{tab:1nf_orig_and_fix}
    \end{subtable}
    \hfill
    \begin{subtable}[ht]{0.50\textwidth}
        \vspace{22px}
        \caption{\textbf{Outputs Before and After Input Data Correction}}
        \label{subtab: correction_1nf}
        \vspace{1px}
        \begin{tabular}{p{0.99\textwidth}}
            {\textbf{T5-base before fix:} Alaska: Dan Sullivan (R) defeated Mark Begich (D) 45.8\% \markerrword{to} \markerrnumber{3.7\%} and Ted Gianoutsos (Independent) 2.0\%, and Virginia: Mark Warner won 49.1\% \markerrword{to} \markerrnumber{2.4\%}.} \\ [1ex]
            {\textbf{T5-base after fix:} Alaska was won by Dan Sullivan (R) with 48.0\% of the vote, and Virginia was won by Mark Warner (D) with 49.1\% of the vote.} \\ [1ex]
            {\textbf{\textsc{Llama 2-7B} before fix: } 2014 United States Senate elections were held on \markerraddition{November 4}, 2014. \markerraddition{The Republican Party gained control of the Senate for the first time since 2006. The Democrats lost their majority in the Senate and Republicans won a net gain of eight seats to take control of the chamber with a 54-46 majority. The Republicans also took control of the House of Representatives, giving them full control of Congress for the first time since 2010.}} \\ [1ex]
            {\textbf{\textsc{Llama 2-7B} after fix:} In the United States Senate elections of 2014, Dan Sullivan (Republican) won in Alaska with 48.0\% of votes against Mark Begich (Democratic) who got 45.8\%. In Virginia, Mark Warner (Democratic) won with 49.1\% of votes against Ed Gillespie (Republican) who got 48.3\%.} \\ [1ex]
        \end{tabular}
        \label{tab:1nf_output}
    \end{subtable}

    
     \vspace{-2px}
     \lineacross{}
     \caption{\textbf{Fixing Multiple leaders' votes that are not atomic, as detailed in \Cref{subsubsec: 1nf}}. `Before fix' shows the errors produced by models with \Cref{tab:1nf_orig} as input and `After fix' shows both models did not produce any factual error with \Cref{tab:1nf_orig_and_fix} as input. The error tokens in outputs are annotated with respective error code in superscript (W for WORD, U for NUMBER and A for ADDITION errors).}
     \label{tab:1nf_problem}
     
\end{table*}

\subsection{\textsc{Llama 2} Models}
In our main study, in addition to T5 models, we included \textsc{Llama 2-7B} (L-7B) and \textsc{Llama 2-13B} (L-13B) models to test our hypothesis on non-standard tabular inputs. We received official model weights from Meta\footnote{We downloaded weights from \url{https://ai.meta.com/resources/models-and-libraries/llama-downloads/} and quantized our models by following the installation instructions from \url{https://github.com/ggerganov/llama.cpp}.} and then quantized our \textsc{Llama 2} models. We ran experiments using the full model weights and the model with 4-bit integer quantization for the ToTTo dataset (\emph{politics} domain). Both produced comparable output quality with no difference in errors for the quantized models. Therefore, we finalized our experiments with 4-bit quantized models to save computational resources and used them for local inference on a MacOS CPU. We set the temperature to 0.0 as it produced fewer errors. Further details on data corrections for these models are discussed in \Cref{sec: experiment_Llama 2_gen}.
\begin{table*}[htbp]
    \lineacross{}
    \setlength\extrarowheight{1pt}
    \customsize

    \definecolor{yellow}{rgb}{1,1,0.5}
    \definecolor{cby}{rgb}{0.91, 0.84, 0.42}
    \definecolor{cbora}{rgb}{1,1,0.5}
    \definecolor{red}{HTML}{DD0000}
        \begin{subtable}[ht]{0.48\textwidth}
        \centering
        \vspace{-18px}
        \subcaption{\textbf{Input Table with Original Cells highlighted in Yellow}}
        \label{tab:result_header_a}
        \vspace{2px}
        \begin{tabular}{P{3cm}| P{2cm}|P{1cm}}
            \multicolumn{3}{l}{\textbf{Page Title:} Minnesota Amendment 2} \\
            \multicolumn{3}{l}{\textbf{Section Title:} Results} \\
            \hline
            \textbf{Choice} & \textbf{Votes} & \textbf{\% Votes} \\
            \hline
            Referendum failed No & 1,588,752 & \cellcolor{yellow}53.84\\
            \hline
            Yes & 1,362,030 &  \cellcolor{yellow}46.16\\
            \hline
        \end{tabular}
       \subcaption{\textbf{Input Table with Corrected Cells highlighted in Yellow}}
       \label{tab:result_header_b}
       \vspace{2px}
       
            \begin{tabular}{P{3cm}| P{2cm}|P{1cm}}
            \hline
            \textbf{Choice} & \textbf{Votes} & \textbf{\% Votes} \\
            \hline
            \cellcolor{cbora}Referendum failed No & 1,588,752 & \cellcolor{yellow}53.84\\
            \hline
            \cellcolor{cbora}Yes & 1,362,030 &  \cellcolor{yellow}46.16\\
        \end{tabular}
        
    \end{subtable}
    \hfill
    \begin{subtable}[ht]{0.50\textwidth}
        \centering
        \vspace{-15px}
        \subcaption{\textbf{Outputs Before and After Input Data Correction}}
        \label{tab:result_header_c}
        \vspace{4px}
        \begin{tabular}{p{0.99\textwidth}}
            \textbf{T5-base before fix}
            The Minnesota Amendment 2 was \markerrword{approved} by 53.84 percent of voters, \markerrcontext{with a turnout of} 46.16 percent.\\
            \vspace{1px}
            \textbf{T5-base after fix:}
            Minnesota Amendment 2 was defeated by 53.84 percent of voters, while 46.16 percent voted in favor of the measure. \\
            \vspace{1px}
            \textbf{\textsc{Llama 2-7B} before fix:}
            53.84\% of Minnesota voters voted \markerrword{for} Amendment 2.\\
            \vspace{1px}

            \textbf{\textsc{Llama 2-7B} after fix:}
            53.84\% of the voters voted no on Amendment 2, while 46.16\% voted yes.\%. \\

        \end{tabular}

     \end{subtable}
     \lineacross{}
     \caption{\textbf{Fixing Insufficient input problems, as detailed in \Cref{subsubsec: insuff}}. `Before' shows WORD and CONTEXT errors due to missing choices of votes. `After' shows these errors are resolved when these cell values are provided.}
     \label{tab:totto_insuf_input_prob}
     
\end{table*}

\subsection{\textbf{Manual Fixes for Input Problems}} 
\label{subsec: fixes}

We followed a systematic procedure to apply manual fixes to specific input problems, as discussed from \Cref{subsubsec: 1nf} to \Cref{subsubsec: order}. A supplementary pseudocode for these fixes is also provided in \Cref{appen: pseudo_code}. The \cref{procedure: main} takes tabular data and title (metadata) from ToTTo as input parameters to execute six functions to gather insights and return with corrected tabular data. The first two functions, \textsc{DataSizeManageableForLongTable} and \textsc{IdentifyLeaderNameOrder} do not correct the tabular input but provide insights on the input problems leading to factual errors in the outputs. The other four functions correct the input problems as shown in \cref{proc:atomic}, \cref{proc:header}, \cref{proc:missing} and \cref{proc:politics}.

\subsubsection{\textbf{Non-Atomic Cell Values}} 
\label{subsubsec: 1nf}
\textbf{Correction:} We corrected the input to store individual leader data, including votes, party, and election details, as a separate column variable. In \Cref{tab:1nf_orig}, when the `candidates' column is not atomic, all models made errors for this tabular data. The corrections involved separating each leader's party and votes into separate columns (see \Cref{tab:1nf_orig_and_fix}). We passed the leading two leaders' data, excluding the remaining leaders, in separate rows for the respective states (Alaska and Virginia). After input correction, T5-b and L-7B outputs in \Cref{tab:1nf_problem} were error-free. While some models omitted other candidate and election details, \textit{all models corrected the previous factual errors after the correction}.

\subsubsection{\textbf{Complex Table Type}} 
\label{subsubsec: complex}

\textbf{Correction:} Complex tabular structures were generally difficult to correct because the election results are in sentence form e.g., see `Results' column in \Cref{tab:complex_table_input_prob} in \Cref{appen: eg_input_problems}. We could only make corrections to the `non-atomic' cell values in `candidates' column. Three models generated `the incumbent senator Coe I. Crawford lost renomination \markerrcontext{to Edwin S. Johnson, a Democrat candidate}.' Losing renomination is when the current senator did not lose the seat to the opponent, but rather failed to be nominated in the primary to stand for reelection. This resulted in \markerrcontext{CONTEXT} error, specifically \textit{when the models made unsupported assumptions about the given input data}, as defined in \Cref{appendix b:anno_procedure}. Even after correction, all models except L-13B made this \markerrcontext{CONTEXT} error (see \Cref{tab:complex_fix_b}). 

\subsubsection{\textbf{Insufficient Input}} 
\label{subsubsec: insuff}
\textbf{Correction:} This is a data annotation problem because it is related to how the ToTTo dataset was created by pairing the sentence description and highlighting only a subset of cells from the Wikipedia table as Input (X). This problem is straightforward to fix by including the missing cells or headers from the tabular data. 

In \Cref{tab:totto_insuf_input_prob}, when the choices for the Minnesota Amendment (i.e., `Referendum failed No' and `Yes' cells) were not included, all models incorrectly generated the favour of votes (\textit{`approved'}, \textit{`for'}) and made an unsupported assumption regarding the \textit{turnout} percentage. As shown in \Cref{tab:totto_insuf_input_prob}, all models corrected the errors after including the vote choices. 

\subsubsection{\textbf{Longer Table Input}} 
\label{subsubsec: longer}
\textbf{Correction:} When dealing with straightforward tabular data, it was easier to correct the input either by \textbf{i.} \textit{correcting the cell values to ensure atomicity} and/or \textbf{ii.} \textit{including the missing cells}. However, this longer tabular input not only had over 100 rows but also included complex structures with nested column headers and row headers which posed difficulties in correcting the input.

Large models such as L-13B and T5-l could process tables with fewer than 20 rows for straightforward inputs. However, T5-b and L-7B struggled to produce factual information even for 20 rows. In our correction procedure, we chose an upper limit of 20 rows and 10 columns to simplify the longer tabular input. A simplified example of this problem with fewer rows is shown in \Cref{tab:long_table} in \Cref{appen: eg_input_problems}.

\subsubsection{\textbf{ToTTo: Politics Specific}} 
\label{subsubsec: politics}
\textbf{Correction:} We included appropriate semantic cues in the headers to help the model differentiate between the generic percentage of votes and the swing percentage of votes. For other errors, we expanded the abbreviation of coalition party names to avoid errors. In \Cref{tab:swing} from \Cref{appen: eg_input_problems}, the original input only had \( \pm \) in the header, resulting in errors from all models including L-13B. After explicitly including `\( \pm \) seats compared to the previous election', both \textsc{Llama 2} models corrected the previous error. However, both fine-tuned T5 models committed mistakes even after this correction. 

\textbf{Specific Semantic Generation Issue:} In some cases, the model cannot determine whether a minister was `shortest-lived' or `longest-lived' only based on their lifespan, birth, and death years from the input. It \textit{may require additional context} to produce accurate text. One such example is shown in \Cref{tab:word_error_totto}. All models with the original data produced incorrect \markerrword{WORD}. After we included the `Total time in office' cell and customized the prompt for \textsc{Llama 2}, both \textsc{Llama 2} models corrected the \markerrword{WORD} error. However, both T5 models did not show any improvement after including this detail. This could be due to the strong influence of the patterns learned from the fine-tuning process for similar tabular structures.

\subsubsection{\textbf{List of Leader Names}} 
\label{subsubsec: order}
\textbf{Correction:} We did not change the order of the leader's name in the title, but we addressed missing vote counts and leader names to map the relations for each record as shown in \Cref{tab:totto_specific_input_prob_order_one}, \Cref{tab:totto_specific_input_prob_order_two}, and \Cref{tab:totto_specific_input_prob_order_three} in the \Cref{appen: eg_input_problems}. 

In \Cref{tab:totto_specific_input_prob_order_two}, the title leader `Chuck DeVore' lost the election among four other party nominees. Both T5 models made errors by focusing on \textit{`Chuck DeVore' as the main candidate}, resulting in \markerrword{WORD} and \markerrcontext{CONTEXT} errors. \markerrnumber{NUMBER} errors are present in all models due to missing votes. After including the votes and party details for all candidates, both \textsc{Llama 2} models corrected all errors. However, both T5 models still made errors stating the title leader either defeated or being defeated by all candidates, possibly due to the \textit{learned patterns during fine-tuning on the ToTTo dataset}.

In \Cref{tab:totto_specific_input_prob_order_three}, the title leader, `Joseph Haslet' ran as a candidate for Governor office in 1804 and 1807 but lost both elections. All models incorrectly generated Haslet won in both years. The correction is only limited to passing the party name and modifying the header from `Subject' to `Candidate'. Despite this change, all models continued to generate errors. Both T5 models struggled to generate the right winning candidate in 5 out of 11 samples (\Cref{fig:t5-base-comparison}) for this input with the same set of headers. 

\subsection{T5 Models on Corrected Data}
\label{subsec:t5_correct}
Based on the insights gathered from different input problems in \cref{sec: input_problems}, we followed the procedure and manually corrected the original tabular input for 210 samples (taken from 124 and 86 `errors' category samples in \Cref{tab:sentence_politics}). We ran both T5-b and T5-l models on the corrected data.

\subsection{\textsc{Llama 2} Models on Corrected Data}
\label{sec: experiment_Llama 2_gen}
From the corrected data, as described in \cref{subsec:t5_correct}, we selected 40 challenging samples from the previous 210 samples as inputs to the \textsc{Llama 2-7B} (L-7B) and \textsc{Llama 2-13B} (L-13B) models. These samples were chosen to cover each of the input problem types described in \cref{sec: input_problems}, guaranteeing a thorough analysis. Within these 40 samples, 21 are associated with general input problems and the remaining 19 are specific to the \emph{politics} input problems, as shown in \Cref{fig:llama2-7b-comparison} and \Cref{fig:llama2-13b-comparison} (in \Cref{appen: eg_input_problems}). 

We studied the \emph{zero-shot} capabilities of L-7B and L-13B models for the 40 challenging samples. For each sample, we employed different prompts tailoring to the content of the tabular input \citep{NEURIPS2022_8bb0d291}. Two common prompts we used are shown in \Cref{appen: prompt_models}. For each sample and each of  L-7B, and L-13B, we examined the outputs before and after correcting the tabular input.

\section{Results and Discussion}
\label{sec: results}

We manually annotated the outputs from all models working on corrected input data, following the same procedure outlined in \Cref{sec: baseline} and defined in \Cref{appendix b:anno_procedure}. In this section, we summarize the results of the total input data corrections and discuss them. The previous section also described and discussed error analysis locally while presenting individual corrections. 

\subsection{\textbf{Error Reductions in T5 Models}}
\label{subsec: t5_results}
The input corrections significantly reduced factual errors, as evident in \Cref{tab:errors_annotated}, which provides a comparison of error counts `before' and `after' input data fixes for each error category. It should be noted that \emph{no prompt or instruction} was provided to these two fine-tuned T5 models.

\begin{table}[htbp]
    \raggedright
    \setlength{\arrayrulewidth}{0.5pt}
    \setlength\extrarowheight{2pt}
    \scriptsize



        \begin{tabular}{p{2.5cm} | P{0.55cm}   P{0.7cm} |  P{0.55cm} P{0.7cm}}
        \toprule
        
        
        
        \multirow{2}{*}{\textbf{Category}} & \multicolumn{2}{c|}{\textbf{T5-base (T5-b)}} & \multicolumn{2}{c}{\textbf{T5-large (T5-l)}} \\  
        & \textbf{Before} & \textbf{After} & \textbf{Before}  & \textbf{After}\\ 
        \midrule

        \textbf{WORD} & 62 & 20 & 31  & 12\\
        \textbf{NAME} & 13 & 3 & 12 & 2 \\
        \textbf{DATE\_DIMENSION} & 7 & 1 & 6  & 1\\
        \textbf{NUMBER} & 12  & 1 & 5 & 0 \\
        \textbf{OTHER} & 5 & 2 & 2 & 0 \\
        \textbf{CONTEXT} & 13 & 3 & 16  & 4 \\
        \textbf{ADDITION} & 2 & 1 & 2  & 1 \\
        \textbf{NON-ENGLISH} & 20 & 20 & 21 & 21 \\ [1ex]
        \midrule
        \textbf{Total errors} & 134 & \textbf{51} & 95 & \textbf{41}\\ 
        \midrule
        \textbf{Error reduction} & \multicolumn{2}{c|}{\textbf{62\%}} & \multicolumn{2}{c}{\textbf{57\%}} \\ [1ex]
        \bottomrule
        \end{tabular}
       

    \caption{\textbf{Count of individual error annotations for 210 samples.} The table compares the error counts between `before' and `after' applying input correction. Each sample can contain multiple errors.}

   
    \label{tab:errors_annotated}
\end{table}
\begin{table*}[t]
    \scriptsize
    \centering
    \setlength{\arrayrulewidth}{0.5pt}
    \setlength\extrarowheight{2pt}

    \begin{adjustbox}{max width=\textwidth}
        
        \begin{tabular}{p{1.75cm} | P{1.2cm}  P{1cm} | P{1.25cm}  P{1cm} | P{1.2cm}  P{1cm} | P{1.2cm}  P{1cm}} 
        \toprule
        \multirow{2}{*}{\textbf{Category}} & \multicolumn{2}{c|}{\textbf{T5-base (T5-b) baseline}} & \multicolumn{2}{c|}{\textbf{T5-large (T5-l) baseline}} & \multicolumn{2}{c|}{\textbf{\textsc{Llama 2-7B} (L-7B)}} & \multicolumn{2}{c}{\textbf{\textsc{Llama 2-13B} (L-13B)}} \\
        & \textbf{Before} & \textbf{After} & \textbf{Before} & \textbf{After }& \textbf{Before}  & \textbf{After} & \textbf{Before}  & \textbf{After}\\ 
        \midrule
        \textbf{WORD} & 27 & 18 & 21 & 12 & 23 & 15 & 22 & 5\\
        
        \textbf{NAME} & 7 & 3 & 6 & 2 & 2 & 1 & 3 & 0\\
        \textbf{DATE\_DIM} & 2  & 1 & 1 & 1 & 2 & 0 & 1 & 0 \\
        \textbf{CONTEXT} & 8  & 3 & 7 & 2 & 6 & 2 & 4 & 0 \\
        \textbf{NUMBER} & 4  & 1 & 1  & 0 & 7 & 1 & 6 & 1 \\
        \textbf{OTHER} & 0 & 0 & 0 & 0 & 5  & 0 & 1 & 0\\
        \textbf{ADDITION} & 0 & 0 & 0 & 0 & 5 & 5 & 7 & 5\\
        \midrule
        \textbf{Total errors} & 48  & \textbf{26} & 36 & \textbf{17} & 50 & \textbf{24} & 45 & \textbf{11}\\ [1ex]
        \midrule
        \textbf{Error \newline reduction(\%)} & \multicolumn{2}{c|}{\textbf{46\%}} & \multicolumn{2}{c|}{\textbf{53\%}} & \multicolumn{2}{c|}{\textbf{52\%}} & \multicolumn{2}{c}{\textbf{76\%}} \\ [1ex]
        \bottomrule
        \end{tabular}
    \end{adjustbox}
    \vspace{2px}
    \caption{\textbf{Individual error count for the 40 challenging samples selected from the previous 210 in \Cref{tab:errors_annotated}.} It shows the comparison of the errors annotated for original data versus corrected data in T5 and \textsc{Llama 2} models.}
    \label{tab:fix_40_samples}
    \vspace{4px}
    \begin{adjustbox}{max width=\textwidth}
        \begin{tabular}{p{1.75cm} | P{1.2cm}  P{1cm} | P{1.25cm}  P{1cm} | P{1.2cm}  P{1cm} | P{1.2cm}  P{1cm}} 
            \toprule
            \multirow{2}{*}{\textbf{Category}} & \multicolumn{2}{c|}{\textbf{T5-base (T5-b) baseline}} & \multicolumn{2}{c|}{\textbf{T5-large (T5-l) baseline}} & \multicolumn{2}{c|}{\textbf{\textsc{Llama 2-7B} (L-7B)}} & \multicolumn{2}{c}{\textbf{\textsc{Llama 2-13B} (L-13B)}} \\
            & \textbf{Before} & \textbf{After} & \textbf{Before} & \textbf{After }& \textbf{Before}  & \textbf{After} & \textbf{Before}  & \textbf{After}\\ 
            \midrule
            \textbf{No error \newline (Higher is better)} & 0  & 12 & 3 & 16 & 2 & 17 & 5  & 23 \\
            \textbf{Omissions} & 0 & 6 & 3 & 8 & 2 & 4 & 4 & 8 \\[1ex]
            \bottomrule
        \end{tabular}
    \end{adjustbox}       
    \vspace{2px}
    \caption{\textbf{Comparison of `No error' and `Omissions' unique count for the same 40 samples}. i. Increase in `No error' count indicates error-free outputs after addressing input problems. ii. Some outputs stopped making factual errors after correction but instead omitted part of the input content, resulting in a higher omission count post-fix.}
    \label{tab:omi_and_err_free}
\end{table*}

\subsection{\textbf{Error Reductions in \textsc{Llama 2} Models}}
\label{subsec: Llama 2_results}
In \Cref{tab:fix_40_samples}, we present the error analysis of 40 difficult samples, validated using L-7B and L-13B models with tailored prompts for each input. Outputs were manually error annotated, and the table compares the error counts before and after input correction. The \emph{L-7B and L-13B} models showed \textbf{reductions of 52\% and 76\% in errors}, after addressing tabular input issues. The \emph{T5-b and T5-l models} demonstrated \textbf{error reductions of 46\% and 53\%} respectively for the 40 difficult samples considered, which was shown for comparison purposes. The insights gathered from the individual error categories after input correction are mentioned below.
\begin{itemize}
    \item \markerrword{WORD} errors, predominantly incorrect verbs and prepositions, were most common among all models. Post-correction, the L-13B model reduced this error category by 77\%, while the T5-b, T5-l, and L-7B models achieved reductions of 33\%, 42\%, and 34\%, respectively.
    \item Input correction led to a reduction in \markerrname{NAME} and \markerrcontext{CONTEXT} errors across all models.
    \item The original input data exhibited more \markerrnumber{NUMBER} errors in \textsc{Llama 2} models compared to T5, which were significantly reduced after correction. Both \textsc{Llama 2} models completely resolved \markerrdate{DATE\_DIMENSION} and \markerrother{OTHER} errors.
    \item \markerraddition{ADDITION} errors are more common in \textsc{Llama 2} models than in T5. Despite including the prompt `Use only the information mentioned in the input table data' and correcting the tabular data, both L-7B and L-13B models still produced five ADDITION errors each.
\end{itemize}

\Cref{tab:omi_and_err_free} shows two rows of data from our analysis of the 40 challenging samples. The first row shows that corrected data leads to an increased number of samples with `no error' (Higher is better). However, the second row shows that omissions have increased after input corrections.
We observed that the models omitted part of the information either when the corrected tabular data had multiple column variables for more than two records or when the tabular structure was complex. In the case of fine-tuned models, it learned to omit some information during the fine-tuning process. In our future work, we plan to study the reasons why this is happening.

\begin{table*}[htbp]
    \centering
    \setlength{\arrayrulewidth}{0.5pt}
    \setlength\extrarowheight{2pt}
    \scriptsize

    {\renewcommand{\arraystretch}{1.5}}

    \begin{adjustbox}{width=\textwidth}
     \begin{tabular}{l|c|c|c|c|c|c|c|c|c|c}
            \toprule
            \textbf{ERROR} & \textbf{ALL AGREE} & \textbf{WORD} & \textbf{NAME} & \textbf{DATE\_DIM} &\textbf{ NUMBER} & \textbf{OTHER} & \textbf{CONTEXT} & \textbf{ADDITION} & \textbf{NO ERROR} & \textbf{TOTAL} \\
            \midrule
            \textbf{WORD} & 17 & 0 & 0 & 0 & 2 & 0 & \textcolor{red}{9} & \textcolor{red}{3} & 0 & 31 \\
            \textbf{NAME} & 3 & 0 & 0 & 0 & 0 & 0 & 2 & 0 & 0 & 5 \\
            \textbf{DATE\_DIM} & 1 & 0 & 0 & 0 & 0 & 0 & 2 & 0 & 0 & 3 \\
            \textbf{NUMBER} & 2 & 2 & 0 & 0 & 0 & 0 & 0 & 0 & 0 & 4 \\
            \textbf{OTHER} & 2 & 0 & 0 & 0 & 0 & 0 & 0 & 0 & 1 & 3 \\
            \textbf{CONTEXT} & 7 & \textcolor{red}{9} & 2 & 2 & 0 & 0 & 0 & \textcolor{red}{6} & 0 & 26 \\
            \textbf{ADDITION} & 12 & 1 & 0 & 0 & 0 & 0 & 5 & 0 & 0 & 18 \\
            \textbf{NO ERROR} & 15 & 0 & 0 & 0 & 0 & 1 & 0 & 0 & 0 & 16 \\
            \bottomrule
    \end{tabular}
    
    \end{adjustbox}
    \begin{tabular}{l}

            {\textbf{Fleiss' kappa overall agreement for three annotators, $k = {(pa - pe)}/{(1 - pe)}$}} = \textbf{0.622} \\

        \end{tabular}
        \lineacross{}
    
        \caption{\textbf{Fleiss` Kappa coefficient: overall agreement on 60 samples among three annotators.} `All agree' column signifies unanimous agreement on error types, while other columns show unique error selections.}
         \label{tab:fleiss}
\end{table*}

\subsection{\textbf{Model-Specific Results for Different Input Problems}}
\label{subsec: results_diff_inputs}

In \Cref{appen: examples}, \Cref{fig:input_problems} presents how the four models are performing before (left bars) and after input corrections (right bars) for each input problem type.  

\textbf{Non-atomic cell values:} T5-l and L-13B models corrected over 90\% of the errors for this problem type, single record and multiple records lacking atomicity (see red and green bars in \Cref{fig:input_problems}, \Cref{appen: eg_input_problems}). T5-b and L-7B models corrected over 60\% of the errors but still struggled with a few samples even after correction. For example, \Cref{fig:t5-large-comparison} shows that T5-large was making more errors for the input problem type labelled `Multiple records lacking atomicity' before correction (in red) and made significant reductions after correction (in green). 

\textbf{Complex table:} Due to the limitations in some tabular inputs, which require additional context, T5-b could not fix the errors. T5-l omitted the error for one sample, while L-7B and L-13B models partially fixed this input problem (see \Cref{fig:input_problems}).

\textbf{Insufficient input:} This data annotation problem fixed all factual errors in T5-b, T5-l and L-13B after correction except for L-7B which added additional information for one sample.

\textbf{Longer table input:} Large models such as L-13B, L-7B and T5-l corrected factual errors for straightforward inputs, especially for tables with fewer than 20 rows. However, T5-b struggled the most to produce factual information. 

\textbf{Politics specific semantic issue:} For the corrected input, L-13B fixed the factual mistakes for 5 out of 8 samples and L-7B fixed factual errors for 4 out of 8 samples. Both T5 models corrected 3 out of 8 samples (see \Cref{fig:input_problems}).

\textbf{List of leader names:} T5 models struggled the most to correct factual errors for this problem. After correction, the L-7B model corrected three samples but produced factual errors for the remaining eight samples. L-13B made factual errors only for two samples (see \Cref{fig:input_problems}). 

While some models struggle with specific inputs, particularly regarding leader name order, tables requiring additional context for complex tabular structures, and semantic issues with symbols, our overall results indicate that correcting tabular inputs improves model outputs.

 
\section{Inter-Annotation Experiment}
\label{sec: inter-anno}

One of the authors manually annotated errors in 1,508 outputs before input correction and 169 problematic outputs after correction (a total of 1,677 from both T5 models). Similarly, the annotation for both \textsc{Llama 2} models covered 160 outputs before and after input correction. 

Two additional annotators annotated 60 outputs each, generated by four different models both before and after input correction. We provided them with a detailed document that included definitions of error categories, guidelines, tabular inputs and output texts for error annotation\footnote{We release our error annotations from our human evaluation on \url{https://github.com/BarkaviSJ/totto_politics_human_annotations}.}. Annotators followed the guidelines and marked the errors. Each annotator spent approximately 3 hours to complete this experiment. 

The annotated errors are shown in a confusion matrix in \Cref{tab:fleiss}, where the `all agree' column shows all annotators agreed on the same error type and other columns show how often each annotator selected a different error type. Although the correct token was usually chosen, disagreements primarily occurred in choosing CONTEXT, ADDITION and WORD error types, as shown in red in \Cref{tab:fleiss}. This might be due to the potential similarities in the definitions of CONTEXT errors and ADDITION errors (see \Cref{appen: categories}). While WORD error is comparatively straightforward, one annotator chose CONTEXT errors instead of WORD errors for a few outputs. Cases where an annotator did not mark any errors were in the minority. The inter-annotation agreement on error category classification for 60 outputs, as shown by a Fleiss' kappa of 0.622, indicates substantial agreement between the three annotators.

\section{Conclusion}
This paper presented a study that quantitatively demonstrates that fixing input problems such as insufficient data and data records containing non-atomic content improves the factual accuracy of output text by as high as 76\% for one of the study models. Correcting inputs also seems to improve the number of entirely error-free output texts. However, we still need to investigate why errors categorised as `omissions' increase after input corrections. In our future work, we aim to explore other tabular datasets for problems with input data and study the generalization of the fixes explored in the current work.

\section*{Limitations}

We acknowledge some limitations in this work. First, we only looked at ToTTo and our scope of corrections to tabular input is limited to errors identified within the \emph{politics} domain of the ToTTo validation set. Second, we did not extend the correction of tabular input for the \emph{politics} domain to the training split of the ToTTo dataset due to the time-consuming process of handling different table headers and metadata. Third, our experiment results are restricted to two specific models (T5 and \textsc{Llama 2}) and may not generalize to other models. 

In our future work, we aim to simplify the definitions 
of the `context' and `addition' error categories, as the annotation experiment revealed disagreement in choosing these error types for some samples, despite annotators marking the same error token.

\section*{Ethics Statement}
This work seeks to address input problems in non-standard tabular structures to reduce factual errors in the output text. We utilized the open-source dataset, ToTTo and maintained the same ground-truth generation as the original dataset. Our input correction did not introduce any further social bias to this dataset. We adopted an error annotation methodology to annotate factual errors and one of the authors performed manual error analysis for this complete study. We sought consent from two additional annotators, the annotators volunteered to participate and annotated errors for 60 output texts each. They had the right to withdraw from the study at any point without facing any consequences. We provided necessary guidelines, instructions and examples for them to annotate errors. 

\section*{Acknowledgements}
We thank Craig Thomson and Adarsa Sivaprasad for their hard work in helping with the annotations in this paper. We thank the anonymous reviewers for their detailed feedback and suggestions which have significantly improved this work. We also thank the NLG (CLAN) reading group at the University of Aberdeen for their invaluable feedback.

\newpage
\appendix
\begin{appendices}
\section{Model fine-tuning specifications}
\label{sec:appendix}
\label{appendix a: model_spec}

The first model, T5-base (T5-b), was fine-tuned on the \textit{full ToTTo training set of 120,761 samples} on a commodity server with a GeForce RTX 2080 GPU. The training took around seven days. The second model, T5-large (T5-l), was fine-tuned on \textit{a subset of 50,000 ToTTo samples} on a secure cloud instance with an NVIDIA A100 GPU, completing in around 48 hours. Both models were fine-tuned using a constant learning rate of 0.0001, with the encoder's maximum length set to 512 tokens and the decoder's maximum length set to 128 tokens for ToTTo's generation task \citep{kale-rastogi-2020-text}. Single-precision floating-point format (FP32) was employed for training on their respective GPU servers. The batch size, beam size and training steps for each model are shown in \Cref{tab:model_spec}.

\begin{figure}[htbp]
    \raggedright
    \setlength{\arrayrulewidth}{0.5pt}
    \setlength\extrarowheight{2pt}
    \footnotesize
    \begin{tabular}{p{1.6cm}|p{1cm} |p{1cm}|p{1.2cm}}
    \toprule
        \textbf{Models} & \textbf{Batch size} & \textbf{Beam Size} & \textbf{Training steps} \\ [0.5ex] 
        \midrule
        T5-base & 2 & 10 & 180,000\\
        T5-large & 4 & 5 & 9,000\\
        \bottomrule
    \end{tabular}
    
    \vspace{0.1px}
    \caption{Model Specifications}
    \label{tab:model_spec}

\end{figure}

\section{Inter-Annotation Procedure for Participants}
\label{appendix b:anno_procedure}
\subsection{Overview}
The Input Data from the ToTTo dataset, includes the Page title, Section Title and a Table with highlighted cells in yellow.  These key parameters are conditionally used for training the neural models to summarize a meaningful and factual Text (as Output) focusing on: (i). highlighted cells in the table (ii). their corresponding header values (iii) The main Title and (iv) The Section Title.

For each of these tabular input data, we provided outputs generated by different neural language models to annotators. Our goal is to evaluate whether the neural outputs remain faithful and produce factually accurate information based on the four parameters from the tabular input. The complete table, including the non-highlighted cells, is provided to offer a clearer understanding of the error annotation task.

\subsection{Domain}
The inputs provided are specific to the domain of \emph{politics}, sourced from Wikipedia tables (as part of the open-domain ToTTo dataset). The political data within these tables is not limited to a single demography. Instead, it encompasses various details from the election processes across multiple countries, including:
\begin{itemize}
    \item Election specifics such as Presidential, state, by-elections, council, district, Legislative Assembly, and other elections unique to particular countries.
    \item Information about Governors, Mayors, Ministers, and Ambassadors (about Foreign Affairs).
    \item Details regarding the Speaker of the Assembly.
\end{itemize}

The first item related to election details is predominantly used in this error annotation task.

\subsection{Error Annotation guidelines}
\label{appen: categories}

We are only interested if the highlighted cell values from the table were used to produce a factually correct sentence. Please also pay attention to the non-highlighted cells in the same row as the highlighted cells, as this might be required in some inputs to generate a meaningful sentence. Other non-highlighted values in the table are not expected to be used for your evaluation. Please read through the output texts and annotate cases for the error categories as mentioned below. Each error is denoted with a superscript for better readability.

\paragraph{\markerrname{NAME}}
\begin{itemize}
    \item When names of the Party, Leader, place (Electorate), Ambassador etc., are wrong (mostly nouns). 
    \item Annotation includes both single tokens or multiple tokens to include the complete names.
    \item Example Output text: \markerrname{Urban Ahlin} is the Deputy Speaker of the Riksdag. Remarks: NAME error because the correct deputy speaker was Tobias Billström as per the tabular data.
    \item Example Output text: Kansas was won by Mitt Romney, Paul Ryan, \markerrname{Barack Obama}, and \markerrname{Joe Biden}, with Romney winning 59.66\% of the popular vote, six electoral votes and 38.05 percent. In this example, Barack Obama and Joe Biden are two NAME errors because they did not win the election.
    \item In general, \markerrname{Wednesday} instead of Tuesday is a NAME error but \markerrdate{May} instead of April is a DATE\_DIMENSION error.
\end{itemize}

\paragraph{\markerrnumber{NUMBER}}
\begin{itemize}
    \item When the number of seats and/or the number of votes and/or \% of votes are incorrect. A single token is marked as an error.
    \item When the A-party won with a majority of 5.5\%. But the correct one is 4.4\%. \markerrnumber{5.5\%} is a NUMBER error.
    \item Output: The voter turnout was 8,90\%, with 10,052 votes. Remarks: The actual turnout was \markerrnumber{81.90\%}. Please note: the error here is NOT with the comma used as decimal (as it is an acceptable decimal operator for international use); Error because the number 81.90 was incorrect.
\end{itemize}

\paragraph{\markerrdate{DATE\_DIMENSION}}
\begin{itemize}
    \item When the Date and/or Month and/or Year are wrong in the generated text, it is annotated as one error.
    \item Example Output text: Cletus Avoka was the Minister for the Interior in the Mills government from 2009 to \markerrdate{2012}. Remarks: 2010 is the right end term of the year.
    \item As a general note, if the Output text did not capture Month and/or Date, but has the correct year, then this is NOT an ERROR. It could go to OMISSION with remarks, Omission of Date and Month.
\end{itemize}

\paragraph{\markerrword{WORD}} 
\begin{itemize}
    \item When incorrect words such as verbs, prepositions, adjectives, adverbs and conjunctions are found in the output. 
    \item Single token is marked as a WORD error in most cases. Multiple tokens are annotated when the auxiliary verb (was), an extension of a prepositional phrase (along with) and others are incorrect.
    \item Example Output text: Carly Fiorina defeated Republican Tom Campbell with 56.4\% of the vote to DeVore's 19.3\%, \markerrword{along with} Al Ramirez and Tim Kalemkarian. Remarks: Fiorina independently defeated all the leaders, so `along with' is wrong.
    \item Example Output text: Ling \markerrword{won} the 2016 senate district against Democrat Josh Newman, with 49.6 percent of the vote. Remarks: Ling lost the election as per the tabular data.
    \item Some of the common WORD errors found in this data are \emph{won, defeated, lost, succeeded}, adjectives such as \emph{current} governor other prepositions (\emph{since, in} and so on). 
\end{itemize}

\paragraph{\markerrcontext{CONTEXT}} 
\begin{itemize}
    \item When the model's output presents information that contradicts or makes \textit{unsupported assumptions about the given input data}. Group of tokens/span of text are annotated as CONTEXT error. 
    \item It can sometimes be tricky to check for this type of error. Please follow the below sequence before marking this error. 
    \begin{itemize}
        \item In case of simple misrepresentation based on the information in the input data, it would be easier to mark the token as NAME, NUMBER, DATE\_DIMENSION or WORD error. 
        \item In the case of a complex table structure, the outputs are likely to mess up completely with the overall information in the provided input data. In this case, it is hard to mark individual errors. Please go ahead and mark the group of tokens/ span of text and annotate it as a CONTEXT error.
        \item For example, the output is: In the 2006 election \markerrcontext{for mayor of Florence Pendleton}, Michael D. Brown received 62,415 votes while Philip Pannell received 21,552 votes and \markerrcontext{write-in candidates} received 1,363 votes. 
        Annotation remarks: 
        \begin{itemize}
            \item \markerrcontext{for mayor of Florence Pendleton} - the name of a person is misrepresented as electorate (jurisdiction) in the output. 
            \item \markerrcontext{write-in candidates} -  this implies there was more than one write-in candidate.
        
        \end{itemize}
    \end{itemize}
\end{itemize}

\paragraph{\markerraddition{ADDITION}}
\begin{itemize}
    \item When the model’s outputs have \textit{added words, phrases, or details} that either diverge from the input's main topic or are unsupported by the given context. 
    \item Single or group of tokens are annotated as ADDITION error.
    \item Example Output text: 2007 Algerian legislative election \markerraddition{was held on May 17, 2007}. The results were as follows: 24 political parties won a total of 389 seats in the National Assembly. \textbf{Remarks:} The date and month are additional information, that are not provided in the input. This is marked as an ADDITION error. Other details in the output are correct. 
\end{itemize}
\paragraph{\markerrother{OTHER}} 
\begin{itemize}
    \item When the output repeats the same input multiple times producing garbage data. In some cases, when the table data has the political party name in the abbreviation, it produces garbage output. For example, when the tabular data has a party name in abbreviation, it tries to produce a strange output. Output text: \markerrother{GSSSDULSVDHSS} gained 5.31\% of the vote
    \item When the output is incomplete for longer table input or when the output repeats the same input multiple times without producing a complete text.
    \item When punctuation symbols are placed in inappropriate places, for example, an apostrophe is missed for the Name of the Leader or Place. 
\end{itemize}
\paragraph{\markerrnoneng{NON-ENGLISH}}
\begin{itemize}
    \item when the Unicode characters in non-English names are either replaced with special characters or when these Unicode characters are omitted.
    \item For example, \markerrnoneng{Pawe Gra} is a member of Sejm. \textbf{Remarks:} Paweł Graś is the correct name here.

\end{itemize}

\section{\textsc{Llama 2} prompts}
\label{appen: prompt_models}

The below prompt is for the corrected tabular data when the input table has party name, candidate and votes.
\begin{quote}
    `Given the input table data, the task is to: \\
    (i). Identify the party name, candidate name, and the number of votes received by each candidate. \\
    (ii). Determine the winner based on the highest number of votes. Then, put together the gathered information from (i) and (ii) into a single coherent sentence. Input table data: <Linearized table data>'
\end{quote}

Below is one of the prompts used for the original table data without mentioning any specific fields.
\begin{quote}

`The task is to summarize the information from the given input table data into a single coherent sentence. Use only the information mentioned in the input table data. Input table data is: <Linearized table data>'
\end{quote}

\section{Steps for Correcting ToTTo Tabular Input}
\label{appen: pseudo_code}

\begin{algorithm*}[!htbp]

\caption{\textbf{Manual Correction Procedure for ToTTo Tabular Input} (elaborated in \Cref{subsec: fixes}). This main function takes tabular data and title details as input parameters and returns the corrected tabular data. In all functions, we excluded row and column indices for simplicity and readability. }
\label{procedure: main}
\begin{algorithmic} 
\Function{MainCorrectionProcedure}{$tabularData, title$}
    \If{\textbf{not} \Call{DataSizeManageableForLongTable}{$tabularData$}}
        \State \Return "Please simplify the tabular data with fewer records", null
    \EndIf
    \State $leaderName, recordedData \gets \Call{IdentifyLeaderNameOrder}{tabularData, title}$
    \State $correctionsMade \gets \textbf{false}$
    \State $correctionsMade \gets \Call{CorrectNonAtomicCells}{tabularData} \textbf{ or } correctionsMade$
    \State $correctionsMade \gets \Call{UpdateHeaders}{tabularData} \textbf{ or } correctionsMade$
    \State $correctionsMade \gets \Call{AddressMissingValues}{tabularData} \textbf{ or } correctionsMade$
    \State $correctionsMade \gets \Call{ReplaceSymbols}{tabularData} \textbf{ or } correctionsMade$
    \If{$correctionsMade$}
        \State \Return $(tabularData$, "Leader Data: " + $recordedData)$ \Comment{Returns corrected input and leader data}
    \Else
        \State \Return $(tabularData$, "Leader Data: " + $recordedData)$ \Comment{Returns original input (if no issues) and leader data}
    \EndIf
\EndFunction
\end{algorithmic}
\end{algorithm*}

\begin{algorithm*}[htbp]
\caption{\textbf{Verify the number of rows and columns in Longer Tabular Input} (discussed in \Cref{subsubsec: longer}). This function validates the maximum allowable number of rows and columns for the tabular data and returns true or false to the main function.}
\label{proc:long}

\begin{algorithmic}
\Function{DataSizeManageableForLongTable}{$tabularData$}
    \State $maxRows \gets 20$ 
    \State $maxCols \gets 10$ 
    \State \Return $(\text{length}(\textit{tabularData}) \leq maxRows) \textbf{ and } (\text{length}(\textit{tabularData}[0]) \leq maxCols)$ 
\EndFunction
\end{algorithmic}
\end{algorithm*}

\begin{algorithm*}[!htbp]
\caption{\textbf{Identify Leader Name Order in Title and Table Rows} (discussed in \Cref{subsubsec: order}). This function verifies three main scenarios for the order of leader names in the input. It returns a tuple with title information for leader name and a list of leader names from tabular input depending on the scenario. Description for each scenario is briefly commented. }
\label{proc:order}
\begin{algorithmic}
\Function{IdentifyLeaderNameOrder}{$\textit{tabularData}, \textit{title}$}
    \State $leaderNameFromTitle \gets \text{ExtractLeaderNameFromTitle}(\textit{title})$ 
    \State $leaderNamesFromRows \gets [\,]$ 
    \For{$\text{each row} \textbf{ in } \textit{tabularData}$} 
        \For{$\text{each cell} \textbf{ in } \text{current row}$} 
            \If{leader\_name is found in cell} 
                \State Add leader\_name to $leaderNamesFromRows$ \Comment{Record leader's names from rows}
            \EndIf
        \EndFor
    \EndFor \\
    \Comment{Scenario 1: Leader's name found in title}
    \If{leaderNameFromTitle is not None} 
        \State $recordedData \gets [\,]$ \Comment{To store sequential order of leader names}
        \State Add $leaderNameFromTitle$ to $recordedData$
        \For{$\text{each leader\_name} \textbf{ in } \textit{leaderNamesFromRows}$} 
            \State Add leader\_name to $recordedData$ 
        \EndFor
        \State \Return ($leaderNameFromTitle, recordedData$) \\
        \hspace{3mm} \Comment{for e.g., this could return ("b", ["a", "b", "c"])} 
    \Else \\
    \Comment{Scenario 2: Leader's name not found in title}
        \If{$\text{length of } leaderNamesFromRows > 0$} \Comment{if leader's name is in at least one row}
            \State $recordedData \gets \text{"Leader name not in title"}$ 
            \For{$\text{each leader\_name} \textbf{ in } \textit{leaderNamesFromRows}$} 
                \State Add leader\_name to $recordedData$ 
            \EndFor
            \State \Return ($leaderNameFromTitle, recordedData$) \\
            \hspace{3mm} \Comment{for e.g., this could return (None, ["Leader name not in title", "a", "b", "c"])}
        \Else \\
        \Comment{Scenario 3: Leader name neither in title nor in rows}
            \State \Return ($leaderNameFromTitle, \text{"Leader name not in table"}$) \\
            \hspace{3mm} \Comment{for e.g., this could return (None, ["Leader name not in table"])}
        \EndIf
    \EndIf
\EndFunction
\end{algorithmic}
\end{algorithm*}

\begin{algorithm*} [!htbp]

\caption{\textbf{Correct Non-Atomic Cells} (discussed in \Cref{subsubsec: 1nf}).  If a non-atomic cell is found, the \texttt{CorrectCell} function separates multiple atomic values into individual columns. We follow our manual correction procedure in \texttt{CellIsNonAtomic} and \texttt{CorrectCell}. The function returns the corrected tabular data along with a flag indicating if any corrections were made.} 
\label{proc:atomic}
\begin{algorithmic}
\Function{CorrectNonAtomicCells}{$tabularData$}
    \State $correctionsMade \gets \textbf{false}$
    \ForAll{$row \textbf{ in } tabularData$}
        \ForAll{$cell \textbf{ in } row$}
            \If{$\text{CellIsNonAtomic}(cell)$}
                \State $correctedCell \gets \text{CorrectCell}(cell)$ \Comment{Separate multiple values into individual columns}
                \State $tabularData[cell] \gets correctedCell$ \Comment{Update the corrected cell value}
                \State $correctionsMade \gets \textbf{true}$
            \EndIf
        \EndFor
    \EndFor
    \State \Return ($tabularData, correctionsMade$)
\EndFunction

\end{algorithmic}
\end{algorithm*}

\begin{algorithm*}[!htbp]
\caption{\textbf{Update Column and Row Headers to Atomic cells} (discussed in \Cref{subsubsec: longer}). For each cell, the tabular input has header\_value as true or false. When the header value is true, we manually verify the function \texttt{HeaderIsIncorrect} and correct the logic in \texttt{UpdateHeader(cell)}. The function returns the corrected tabular data with corrections made flag.}
\label{proc:header}
\begin{algorithmic}
\Function{CorrectHeaders}{$tabularData$}
    \State $correctionsMade \gets \textbf{false}$
    \ForAll{$row \textbf{ in } tabularData$}
        \ForAll{$cell \textbf{ in } row$}
            \If{$\text{cell.header\_value} = \textbf{true}$} 
                \If{$\text{HeaderIsIncorrect}(cell)$}
                    \State $correctedHeader \gets \text{UpdateHeader}(cell)$ \Comment{Update column and row headers}
                    \State $tabularData[cell] \gets correctedHeader$ \Comment{Update the corrected header}
                    \State $correctionsMade \gets \textbf{true}$
                \EndIf
            \EndIf
        \EndFor
    \EndFor
    \State \Return $(tabularData, correctionsMade)$
\EndFunction

\end{algorithmic}
\end{algorithm*}

\begin{algorithm*}[!htbp]
\caption{\textbf{Address Missing cell Values} (discussed in \Cref{subsubsec: insuff}). For each row, we verify the missing cell values in \texttt{RowHasMissingValues} and pass the right cell values in \texttt{FillCellValues} through a manual process. The function returns the corrected tabular data with corrections made flag.}
\label{proc:missing}
\begin{algorithmic} 
\Function{AddressMissingValues}{$tabularData$}
    \State $correctionsMade \gets \textbf{false}$
    \ForAll{$row \textbf{ in } tabularData$}
        \If{$\text{RowHasMissingValues}(row)$}
        \State $correctedRow \gets \text{FillCellValues}(row)$  \Comment{Pass missing cells in the row}
            \State ${tabularData}[row] \gets correctedRow$
            \State $correctionsMade \gets \textbf{true}$
        \EndIf
    \EndFor
    \State \Return $(tabularData, correctionsMade)$
\EndFunction
\end{algorithmic}
\end{algorithm*}

\begin{algorithm*}[!htbp]
\caption{Replace Politics Domain-Specific Symbols and Abbreviate Party Names with Correct Semantics/words (discussed in \Cref{subsubsec: politics}). For all cells in the table, we verify \texttt{containsDomainSpecificSymbols} and \texttt{containsPartyAbbreviations} functions and correct the values using \texttt{substituteSymbolWithEquivalent} and \texttt{containsPartyAbbreviations} through a manual process. The function returns the corrected tabular data with corrections made flag.}
\label{proc:politics}
\begin{algorithmic}
\Function{replaceSymbols}{\textit{tabularData}}
    \State $correctionsMade \gets \textbf{false}$
    \ForAll{$cell \textbf{ in } \textit{tabularData}$}
        \If{$\text{containsDomainSpecificSymbols}(cell)$}
            \State $correctedValue \gets \text{substituteSymbolWithEquivalent}(cell)$ \Comment{Replace symbols with words}
            \State $tabularData[cell] \gets correctedValue$ \Comment{Update the corrected cell value}
            \State $correctionsMade \gets \textbf{true}$
        \ElsIf{$\text{containsPartyAbbreviations}(cell)$}
            \State $correctedValue \gets \text{abbreviatePartyNames}(cell)$ \Comment{Abbreviate party names}
            \State $\textit{tabularData}[cell] \gets corrected value$ \Comment{Update the corrected cell value}
            \State $correctionsMade \gets \textbf{true}$
        \EndIf
    \EndFor
    \State \Return $(tabularData, correctionsMade)$
\EndFunction
\end{algorithmic}
\end{algorithm*}

In addition to the correction procedure detailed in \Cref{subsec: fixes}, which provides examples of the corrections applied to each tabular input problems, we present a supplementary pseudocode in this section. In \cref{procedure: main}, we have a generic function that takes tabular data and title (metadata) from ToTTo as input parameters. This main algorithm executes six different functions. 

The first two functions, \textsc{DataSizeManageableForLongTable} and \textsc{IdentifyLeaderNameOrder} provide insights on the input problems leading to factual errors in the outputs but do not correct the tabular input. The other four functions, namely \textsc{CorrectNonAtomicCells}, \textsc{UpdateHeaders}, \textsc{AddressMissingValues}, and \textsc{ReplaceSymbols} correct the input problems as presented in \cref{proc:atomic}, \cref{proc:header}, \cref{proc:missing}, and \cref{proc:politics}. We provide brief descriptions for each algorithm in the caption and comments.

\section{Input Problem Types for four models}
\label {appen: examples} 

\Cref{fig:input_problems} shows how each of the four models is performing before and after data corrections for each of the problem types. The left bars for each input problem type represent output scores before data correction, while the right bars represent scores after data correction. Colour coding of output scores helps to identify and understand errors: green indicates no error, red indicates an error, and yellow indicates an omission. 

The number of samples for each model differs because we focus on outputs only when the model had `errors' or `omissions' before correction. This approach emphasizes the actual improvement in corrections. For example, T5-large had errors in 37 samples before correction (left bar), the after-correction (right bar) also shows improvements for the same 37 samples.

\input{charts/four_charts}

\section{Input Data Corrections and Output Text}
\label{appen: eg_input_problems}
In this section, the input tabular data with outputs supplements the examples elaborated in \Cref{subsec: fixes} for each of the input problems as presented from \Cref{tab:swing} to \Cref{tab:long_table}.

\begin{table*}[!htbp]
    \lineacross{}
    \setlength\extrarowheight{1pt}

    \definecolor{yellow}{rgb}{1,1,0.5}
    \definecolor{cby}{rgb}{0.91, 0.84, 0.42}
    \definecolor{cbora}{rgb}{1,1,0.5}
    \definecolor{red}{HTML}{DD0000}
    \vspace{-10px}

            \begin{subtable}[ht]{0.47\textwidth}
            \centering
            \footnotesize
            \vspace{-10px}
            \captionsetup{justification=raggedright,singlelinecheck=false, font=footnotesize}
            \caption{\textbf{Input Table with Original Cells highlighted in Yellow}}
            \vspace{4px}
            \begin{tabular}{p{2.2cm}|p{0.6cm}|p{1.5cm}}
                               
                \multicolumn{3}{p{7cm}}{\textbf{Page Title:} 2012 Cardiff South and Penarth by-election} \\
                \multicolumn{3}{p{6cm}}{\textbf{Section Title:} By-election result} \\
                \hline
                \textbf{Party} & \textbf{\%} & {\textbf{\(\pm \)}} \\
                \hline
                Conservative & 19.9 & 8.4 \\
                \hline
                \cellcolor{yellow}Liberal Democrat & 10.8 & \cellcolor{yellow}-11.5 \\ [1ex]
                \hline

            \end{tabular}
           
           \label{tab:swing_a}
       \end{subtable}
       \begin{subtable}[ht]{0.50\textwidth}
            \centering
            \footnotesize
            \vspace{-7px}
            \captionsetup{justification=raggedright,singlelinecheck=false, font=footnotesize}
            \caption{\textbf{Input Table with Corrected Cells highlighted in Yellow}}
            \vspace{1px}
           \begin{tabular}{p{2.2cm}|p{0.5cm}|p{4cm}}
                               
                \multicolumn{3}{p{7cm}}{\textbf{Page Title:} 2012 Cardiff South and Penarth by-election} \\
                \multicolumn{3}{p{7cm}}{\textbf{Section Title:} By-election result} \\
                \hline
                \textbf{Party} & \textbf{\%} & \cellcolor{cbora}{\textbf{\(\pm \) \% difference with previous election}} \\
                \hline
                Conservative & 19.9 & 8.4 \\
                \hline
                \cellcolor{yellow}Liberal Democrat & 10.8 & \cellcolor{yellow}-11.5 \\ [1ex]
                \hline

            \end{tabular}
       
        \label{tab:swing_b}
        \end{subtable}
        \hfill
        \begin{subtable}[ht]{\textwidth}
        \centering
        \footnotesize
        \captionsetup{justification=raggedright,singlelinecheck=false, font=footnotesize}
        \caption{\textbf{Outputs Before and After Input Data Correction}}
        \vspace{2px}
        \begin{tabular}{p{0.99\textwidth}}
            {\textbf{T5-large before fix:} In the 2012 Cardiff South and Penarth by-election, the Liberal Democrats \markerrword{won} 11.5\% of the vote.}\\ 
            \vspace{1px} 
            {\textbf{T5-large after fix:} In the 2012 Cardiff South and Penarth by-election, the Liberal Democrats \markerrword{lost by} 11.5\%.} \\ 
            \vspace{1px} 
            {\textbf{\textsc{Llama 2-13B} before fix:} 2012 Cardiff South and Penarth by-election was \markerrword{won} by the Liberal Democrats with a \markerrword{margin} of -11.5\%}. \\ 
            \vspace{1px} 
            {\textbf{\textsc{Llama 2-13B} after fix:} 2012 Cardiff South and Penarth by-election saw Liberal Democrat candidate lose by 11.5\% of the vote compared to the previous election.} \\ [1ex]

        \end{tabular}
        
        \end{subtable}
     \lineacross{}
     \caption{\textbf{ToTTo specific: Swing percentage difference with previous election, as detailed in \Cref{subsubsec: politics}.} Before fix represents the output generated from input \Cref{tab:swing_a} and after fix represents the output generated after explicitly changing the header, as shown in input \Cref{tab:swing_b}.}
    \label{tab:swing}

\end{table*}

\begin{table*}[htbp]
    \lineacross{}
    \setlength\extrarowheight{2pt}
    \footnotesize

    \definecolor{yellow}{rgb}{1,1,0.5}
    \definecolor{cby}{rgb}{0.91, 0.84, 0.42}
    \definecolor{cbora}{rgb}{1,1,0.5}
    \definecolor{red}{HTML}{DD0000}
    \vspace{-15px}
        \begin{subtable}[ht]{1.0\textwidth}
        \centering
        \vspace{5px}
        \captionsetup{justification=raggedright,singlelinecheck=false, font=footnotesize}
        \caption{\textbf{Input Table with Original Cells highlighted in Yellow}}
        \vspace{5px}
        \begin{tabular}{P{1cm}|P{1.4cm}|P{1.5cm}| P{3.8cm}|P{5cm}}
            \multicolumn{5}{l}{\textbf{Page Title:} 1914 United States Senate elections} \\
            \multicolumn{5}{l}{\textbf{Section Title:} Races leading to the 64th Congress} \\
            \hline
            \multirow{2}{*}{\textbf{State}} & \multicolumn{2}{c|}{\textbf{Incumbent}} & \multirow{2}{*}{\textbf{Results}} & \multirow{2}{*}{\textbf{Candidates}} \\
            \cline{2-3}
             & \textbf{Senator} & \textbf{Party} & & \\
            \hline
            \cellcolor{yellow}South Dakota & \cellcolor{yellow}Coe I.Crawford & \cellcolor{yellow}Republican & \cellcolor{yellow}Incumbent lost renomination. New senator elected. Democratic gain. & \cellcolor{yellow}Edwin S. Johnson (Democratic) 48.32\% Charles H. Burke (Republican) 44.47\% \\
            \hline
        \end{tabular}
       
       \label{tab:complex_a}
    \end{subtable}
    \hfill
    \begin{subtable}[ht]{1.0\textwidth}
        \centering
        \captionsetup{justification=raggedright,singlelinecheck=false, font=footnotesize}
        \caption{\textbf{Outputs with Original Input before Input Data Correction}}
        \begin{tabular}{p{0.99\textwidth}}
            \textbf{T5-base:}
            In South Dakota, Republican Coe I. Crawford lost renomination \markerrcontext{to Democrat Edwin S. Johnson}.\\
            \textbf{T5-large:} South Dakota: Republican incumbent Coe I. Crawford lost his bid for the seat \markerrcontext{to Democratic challenger Edwin S. Johnson}. \\
            \textbf{\textsc{Llama 2-7B}:} 1914 United States Senate elections in South Dakota were held on \markerraddition{November 3}, 1914. Incumbent Republican Senator Coe I. Crawford lost renomination \markerrcontext{to Democratic candidate Edwin S. Johnson}. Johnson defeated \markerrname{Charles H. Dunnigan} by \markerrword{a margin of} 48.32\% to \markerrnumber{47.56\%}. This was a Democratic gain from the previous election.\\
            \textbf{\textsc{Llama 2-13B}:} 1914 United States Senate elections were held on \markerraddition{November 3}, 1914 to elect members of the United States Senate in various states. The Republican Party lost one seat and the Democratic Party gained one seat.

        \end{tabular}
        
        
        \label{tab:complex_b}
     \end{subtable}
     \begin{subtable}[ht]{1.0\textwidth}
        \centering
        \vspace{5px}
        \captionsetup{justification=raggedright,singlelinecheck=false, font=footnotesize}
        \caption{\textbf{Input Table with Corrected Cells highlighted in Yellow}}
        \vspace{5px}
        \begin{tabular}{P{1cm}|P{1.7cm}|P{1.7cm}| P{3cm}|P{1.5cm}|P{1.5cm}|P{1.5cm}}
            \multicolumn{6}{l}{\textbf{Page Title:} 1914 United States Senate elections} \\
            \multicolumn{6}{l}{\textbf{Section Title:} Races leading to the 64th Congress} \\
            \hline
            {\textbf{State}} & {\textbf{Incumbent Senator}} & \textbf{Incumbent Party} & {\textbf{Results}} & {\textbf{Candidate Senator}} & {\textbf{Candidate Party}} & {\textbf{Candidate votes \%}}\\
            \hline
            \cellcolor{yellow}South Dakota & \cellcolor{yellow}Coe I.Crawford & \cellcolor{yellow}Republican & \cellcolor{yellow}Incumbent lost renomination. New senator elected. Democratic gain. & \cellcolor{cbora}Edwin S. Johnson & \cellcolor{cbora}Democratic & \cellcolor{cbora}48.32\% \\
            \hline
            South Dakota & Coe I.Crawford & Republican & Incumbent lost renomination. New senator elected. Democratic gain. & \cellcolor{cbora}Charles H. Burke & \cellcolor{cbora}Republican & \cellcolor{cbora}44.47\% \\
            \hline
        \end{tabular}
       
       \label{tab:complex_fix_a}
    \end{subtable}
    \hfill
    \begin{subtable}[ht]{1.0\textwidth}
        \vspace{5px}
        \captionsetup{justification=raggedright,singlelinecheck=false, font=footnotesize}
        \subcaption{\textbf{Outputs After Input Data Correction}}
        \begin{tabular}{p{0.99\textwidth}}
        \textbf{T5-base:}
            In South Dakota, Republican Coe I. Crawford lost renomination \markerrcontext{to Democrat Edwin S. Johnson}.\\
            \textbf{T5-large:} South Dakota: Republican incumbent Coe I. Crawford lost his bid for the seat \markerrcontext{to Democratic challenger Edwin S. Johnson}. \\
            \textbf{\textsc{Llama 2-7B}:}
            1914 United States Senate elections were held on \markerraddition{November 3,} 1914. In South Dakota, Coe I. Crawford, the Republican incumbent Senator lost renomination \markerrcontext{to Edwin S. Johnson}, a Democratic candidate. Charles H., the Republican candidate also ran for the seat but he was defeated by Johnson. Therefore, it is a Democratic gain in this election. \\
            \textbf{\textsc{Llama 2-13B}:} 1914 United States Senate elections were held on \markerraddition{November 3}, 1914 to elect members of the United States Senate in various states. Races leading to the 64th Congress included South Dakota where Coe I. Crawford (Republican) lost renomination and Edwin S. Johnson (Democratic) was elected.
        \end{tabular}
        \label{tab:complex_fix_b}
    \end{subtable}
    \lineacross{}
    \caption{\textbf{Complex Table input problems, as detailed in \Cref{subsubsec: complex}}. Difficult to fix this complex input type, yet \textsc{Llama 2-13B} produced most of the facts right except for ADDITION error. Lost renomination `to Democrat candidate' is a CONTEXT error, because the model made unsupported assumptions about the given input. Losing renomination is when the current senator failed to be nominated in the primary to stand for reelection.}
    \label{tab:complex_table_input_prob}
    
\end{table*}

\begin{table*}[htbp]
    \vspace{-10px}
    \lineacross{}
    \setlength\extrarowheight{2pt}

    \definecolor{yellow}{rgb}{1,1,0.5}
    \definecolor{cby}{rgb}{0.91, 0.84, 0.42}
    \definecolor{cbora}{rgb}{1,1,0.5}
    \definecolor{red}{HTML}{DD0000}

     \begin{subtable}[ht]{0.48\textwidth}
            \centering
            \footnotesize
            \vspace{-16px}
            \captionsetup{justification=raggedright,singlelinecheck=false, font=footnotesize}
            \caption{\textbf{Input Table with Original Cells highlighted in Yellow}}
            \vspace{1px}
            \begin{tabular}{p{2.1cm}| p{1.4cm}|p{1cm}|p{1.1cm}}
            \multicolumn{4}{p{7.3cm}}{\textbf{Page Title:} List of Japanese prime ministers by longevity} \\
            \multicolumn{4}{p{7cm}}{\textbf{Section Title:} Longevity of Japanese prime ministers} \\
            \hline
            \textbf{Prime Minister} & \textbf{Total time in Office} & \textbf{Date of Death} & \textbf{Lifespan}\\
            \hline
            \cellcolor{yellow} Sanetomi Sanjō & 60 days & 28 Feb 1891 & \cellcolor{yellow} 53 years, 352 days\\
            \hline
    
            \end{tabular}
           
           \label{tab:word_a}
       \end{subtable}
       \begin{subtable}[ht]{0.50\textwidth}
            \centering
            \footnotesize
            \vspace{-16px}
            \captionsetup{justification=raggedright,singlelinecheck=false, font=footnotesize}
            \caption{\textbf{Input Table with Corrected Cells highlighted in Yellow}}
            \vspace{1px}
            \begin{tabular}{p{2.1cm}| p{1.4cm}|p{1cm}|p{1.1cm}}
            \multicolumn{4}{p{7.3cm}}{\textbf{Page Title:} List of Japanese prime ministers by longevity} \\
            \multicolumn{4}{p{7cm}}{\textbf{Section Title:} Longevity of Japanese prime ministers} \\
            \hline
            \textbf{Prime Minister} & \textbf{Total time in Office} & \textbf{Date of Death} & \textbf{Lifespan}\\
            \hline
            \cellcolor{yellow} Sanetomi Sanjō & \cellcolor{cbora}60 days & 28 Feb 1891 & \cellcolor{yellow} 53 years, 352 days\\
            \hline
    
            \end{tabular}
           
            \label{tab:word_b}
            \end{subtable}
            \hfill
            \begin{subtable}[ht]{\textwidth}
                \centering
                \footnotesize
                \captionsetup{justification=raggedright,singlelinecheck=false, font=footnotesize}
                \caption{\textbf{Outputs Before and After Input Data Correction}}
                \vspace{2px}
                \begin{tabular}{p{0.99\textwidth}}
                    \textbf{T5-base and T5-large before and after fix (no improvement):} \markerrnoneng{Sanj Sanetomi} was the \markerrword{longest-lived} prime minister in Japan at 53 years, 352 days.\\
                    \vspace{1px}
                    \textbf{\textsc{Llama 2-7B} before fix:} The \markerrword{longest-serving} Japanese prime minister was Sanjō Sanetomi who served for 53 years and 352 days.\\
                    \textbf{\textsc{Llama 2-7B} after data fix:} Sanjō Sanetomi was the Japanese prime minister and served for a total of 60 days in office. He died at age 53 years and 352 days. \\
                    \vspace{1px}
                    \textbf{\textsc{Llama 2-13B} before data fix:} The \markerrword{longest-lived} Japanese prime minister was Sanjō Sanetomi, who served for 53 years and 352 days.\\
                    \textbf{\textsc{Llama 2-13B} after data fix:} Sanjō Sanetomi was the prime minister of Japan for a total of 60 days. He lived for 53 years and 352 days.\\
                \end{tabular}
                
        \end{subtable}
    \lineacross{}
    \caption{\textbf{ToTTo specific - WORD error made by all models before fix, as detailed in \Cref{subsubsec: politics}.} After including `Total time in Office', \textsc{Llama 2} models fixed the `longest-lived' word error and introduced other addition errors. We then revised the prompt and fixed the addition errors. \textit{T5 models had no improvements after including information.} }

    \label{tab:word_error_totto}
    
\end{table*}

\begin{table*}[htbp]
    \lineacross{}
    \setlength\extrarowheight{1pt}

    \definecolor{yellow}{rgb}{1,1,0.5}
    \definecolor{cby}{rgb}{0.91, 0.84, 0.42}
    \definecolor{cbora}{rgb}{1,1,0.5}
    \definecolor{red}{HTML}{DD0000}

        \begin{subtable}[ht]{0.48\textwidth}
            \centering
            \footnotesize
            \vspace{-16px}
            \captionsetup{justification=raggedright,singlelinecheck=false, font=footnotesize}
            \caption{\textbf{Input Table with Original Cells highlighted in Yellow}}
            \vspace{1px}
            \begin{tabular}{P{2cm}| P{2.5cm}|P{1.2cm}}
                \multicolumn{3}{l}{\textbf{Page Title:} Ling Ling Chang} \\
                \multicolumn{3}{l}{\textbf{Section Title:} 2016 Senate run} \\
                \hline
                \multicolumn{3}{l}{\textbf{California's 29th State Senate district election}} \\
                \hline
                \textbf{Party} & \textbf{Candidate} & \textbf{\% Votes} \\
                \hline
                \cellcolor{yellow}Democratic & \cellcolor{yellow}Josh Newman & 50.4\\
                \hline
                Republican & \cellcolor{yellow}Ling Ling Chang & \cellcolor{yellow}49.6\\
                \hline
            \end{tabular}
           
           \label{tab:result_one_a}
       \end{subtable}
       \begin{subtable}[ht]{0.50\textwidth}
            \centering
            \footnotesize
            \vspace{-16px}
            \captionsetup{justification=raggedright,singlelinecheck=false, font=footnotesize}
            \caption{\textbf{Input Table with Corrected Cells highlighted in Yellow}}
            \vspace{1px}
           \begin{tabular}{P{2cm}| P{2.5cm}|P{1.2cm}}
                \multicolumn{3}{l}{\textbf{Page Title:} Ling Ling Chang} \\
                \multicolumn{3}{l}{\textbf{Section Title:} 2016 Senate run} \\
                \hline
                \multicolumn{3}{l}{\textbf{California's 29th State Senate district election}} \\
                \hline
                \textbf{Party} & \textbf{Candidate} & \textbf{\% Votes} \\
                \hline
                \cellcolor{yellow}Democratic & \cellcolor{yellow}Josh Newman & \cellcolor{cbora}50.4\\
                \hline
                \cellcolor{cbora}Republican & \cellcolor{yellow}Ling Ling Chang & \cellcolor{yellow}49.6\\
                \hline
            \end{tabular}
       
        \label{tab:result_one_b}
        \end{subtable}
        \hfill
        \begin{subtable}[ht]{\textwidth}
        \centering
        \footnotesize
        \captionsetup{justification=raggedright,singlelinecheck=false, font=footnotesize}
        \caption{\textbf{Outputs Before and After Input Data Correction}}
        \vspace{2px}
        \begin{tabular}{p{0.99\textwidth}}
            \textbf{T5-base and T5-large before fix:}
            Chang \markerrword{defeated} Democrat Josh Newman with 49.6\% of the vote.\\
            \vspace{0.1px}
            \textbf{T5-base and T5-large after fix:}
            Ling Ling Chang was defeated by Josh Newman, a Democrat, with 50.4\% of the vote to Chang's 49.6\%. \\
            \vspace{0.1px}
            \textbf{\textsc{Llama 2-13B} before fix:}
            Ling Ling Chang \markerrword{won} the 2016 Senate run with 49.6\% of the vote.\\
            \vspace{0.1px}

            \textbf{\textsc{Llama 2-13B} after fix:}
            Ling Ling Chang, who was running for the Senate seat in California, received 49.6\% of votes while Josh Newman, her opponent, received 50.4\%. \\
        \end{tabular}
        
        \end{subtable}
     \lineacross{}
     \caption{\textbf{Fixing ToTTo specific input problems: List of leader names for two leaders}. `Before' shows WORD errors due to missing vote percentages and party names for a leader. `After' shows these errors are resolved by all models when the missing values are provided. Key point is T5 model did not produce error even though the title leader `Ling Ling Chang' lost the election.}
     \label{tab:totto_specific_input_prob_order_one}
     
\end{table*}
\begin{table*}[ht]
    \lineacross{}
    \setlength\extrarowheight{1pt}

    \definecolor{yellow}{rgb}{1,1,0.5}
    \definecolor{cby}{rgb}{0.91, 0.84, 0.42}
    \definecolor{cbora}{rgb}{1,1,0.5}
    \definecolor{red}{HTML}{DD0000}
     \begin{subtable}[ht]{0.48\textwidth}
        \centering
        \footnotesize
        \vspace{-14px}
        \captionsetup{justification=raggedright,singlelinecheck=false, font=footnotesize}
        \caption{\textbf{Input Table with Original Cells highlighted in Yellow}}
        \vspace{4px}
        \begin{tabular}{p{1.5cm}| p{2.2cm}|p{1.1cm}|p{0.6cm}}
        \multicolumn{4}{p{6.5cm}}{\textbf{Page Title:} Chuck DeVore} \\
        \multicolumn{4}{p{6.5cm}}{\textbf{Section Title:} As U.S. Senate Candidate} \\
        \hline
        \textbf{Party} & \textbf{Candidate} & \textbf{Votes} & \textbf{\%}\\
        \hline
        Republican & \cellcolor{yellow}Carly Fiorina & 1,315,429 & \cellcolor{yellow}56.4\\
        Republican & \cellcolor{yellow}Tom Campbell & 504,289 & 21.7\\
        \cellcolor{yellow}Republican & \cellcolor{yellow}Chuck DeVore & 452,577 & \cellcolor{yellow}19.3\\
        Republican & \cellcolor{yellow}Al Ramirez & 42,149 & 1.8\\
        Republican & \cellcolor{yellow}Tim Kalemkarian & 19,598 & 0.8\\
        \hline

        \end{tabular}
    \end{subtable}
     \begin{subtable}[ht]{0.50\textwidth}
        \centering
        \footnotesize
        \vspace{-14px}
        \captionsetup{justification=raggedright,singlelinecheck=false, font=footnotesize}
        \caption{\textbf{Input Table with Corrected Cells highlighted in Yellow}}
        \vspace{4px}
        \begin{tabular}{p{1.5cm}| p{2.2cm}|p{1.1cm}|p{0.6cm}}
        \multicolumn{4}{p{6.5cm}}{\textbf{Page Title:} Chuck DeVore} \\
        \multicolumn{4}{p{6.5cm}}{\textbf{Section Title:} As U.S. Senate Candidate} \\
        \hline
        \textbf{Party} & \textbf{Candidate} & \textbf{Votes} & \textbf{\%}\\
        \hline
        \cellcolor{cbora}Republican & \cellcolor{yellow}Carly Fiorina & 1,315,429 & \cellcolor{yellow} 56.4\\
        \cellcolor{cbora}Republican & \cellcolor{yellow}Tom Campbell & 504,289 & \cellcolor{cbora} 21.7\\
        \cellcolor{yellow}Republican & \cellcolor{yellow}Chuck DeVore & 452,577 & \cellcolor{yellow} 19.3\\
        \cellcolor{cbora}Republican & \cellcolor{yellow}Al Ramirez & 42,149 & \cellcolor{cbora}1.8\\
        \cellcolor{cbora}Republican & \cellcolor{yellow}Tim Kalemkarian & 19,598 & \cellcolor{cbora}0.8\\
        \hline

        \end{tabular}
    \end{subtable}
    \hfill
    \begin{subtable}[ht]{0.99\textwidth}
        \footnotesize
        \captionsetup{justification=raggedright,singlelinecheck=false, font=footnotesize}
        \vspace{2px}
        \caption{\textbf{Outputs Before and After Input Data Correction}}
        \vspace{2px}

        \begin{tabular}{p{0.99\textwidth}}


            \textbf{T5-base before fix:} 
            DeVore \markerrword{defeated} Republican Tom Campbell with \markerrnumber{56.4\%} of the vote, \markerrcontext{Al Ramirez with 19.3\%, and Tim Kalemkarian with 19.3\%}.\\
            \textbf{T5-base after fix:} 
            DeVore \markerrword{defeated} Republican nominees Carly Fiorina (56.4\%)\markerrcontext{, Tom Campbell (21.7\%), Al Ramirez (1.8\%) and Tim Kalemkarian (0.8\%)}.\\
            \vspace{1px}
            \textbf{T5-large before fix:} 
            DeVore was defeated by Republican Tom Campbell \markerrnumber{(56.4\%)}, \markerrcontext{Al Ramirez (19.3\%), and Tim Kalemkarian (21.7\%)}.\\
            \textbf{T5-large after fix:} 
            DeVore was defeated in the U.S. Senate race by Republican nominees Carly Fiorina (56.4\%), Tom Campbell (21.7\%)\markerrcontext{, Al Ramirez (1.8\%) and Tim Kalemkarian (0.8\%).} \\
            \vspace{1px}
            \textbf{\textsc{Llama 2-7B} before fix:} Chuck DeVore is a Republican candidate for the U.S. Senate in California. He received 19.3\% of the vote in the primary election, which was held on June 8th, 2010. The other candidates were Carly Fiorina (56.4\%), Tom Campbell \markerrnumber{(24.1\%)} and Al Ramirez \markerrnumber{(10.2\%)}. Tim Kalemkarian \markerrcontext{did not run for this office}.\\
            \textbf{\textsc{Llama 2-7B} after fix:}
            The number of votes for each leader is as follows: Carly Fiorina - 56.4\% Tom Campbell - 21.7\% Chuck DeVore - 19.3\% Al Ramirez - 1.8\% Tim Kalemkarian - 0.8\%. Carly Fiorina won the Republican primary for U.S. Senate in California with 56.4 percent of the vote, followed by Tom Campbell at 21.7 percent and Chuck DeVore at 19.3 percent. \\
            \vspace{1px}
            \textbf{\textsc{Llama 2-13B} before fix:} 
            Carly Fiorina is the leading candidate for U.S. Senate with 56.4\% of the vote, followed by Tom Campbell with \markerrnumber{20.3\%}, Chuck DeVore with 19.3\%, and Al Ramirez with \markerrnumber{4.0\%}.\\
            \textbf{\textsc{Llama 2-13B} after fix:}
            The Republican Party candidate who won the most votes in the U.S. Senate election was Carly Fiorina with 56.4\% of the total votes.\\

        \end{tabular}

     \end{subtable}
    \lineacross{}

    \caption{\textbf{ToTTo specific input problems: List of Leader names for five leaders, as detailed in \Cref{subsubsec: order}.} All models swapped the defeated leader names and/or \% of votes, resulting in WORD, NUMBER and CONTEXT errors. After including the party names and \% of votes for all leaders, \textsc{Llama 2} models corrected the errors. Fine-tuned T5 models still made errors. This could be because of the pattern learned from the fine-tuned training data of ToTTo.}

    \label{tab:totto_specific_input_prob_order_two}

\end{table*}

\begin{table*}[htbp]
    \lineacross{}
    \setlength\extrarowheight{2pt}

    \definecolor{yellow}{rgb}{1,1,0.5}
    \definecolor{cby}{rgb}{0.91, 0.84, 0.42}
    \definecolor{cbora}{rgb}{1,1,0.5}
    \definecolor{red}{HTML}{DD0000}

     \begin{subtable}[ht]{\textwidth}
        \centering
        \footnotesize
        \vspace{-16px}
        \captionsetup{justification=raggedright,singlelinecheck=false, font=footnotesize}
        \caption{\textbf{Input Table with Original Cells highlighted in Yellow}}
       \vspace{1px}
        \begin{tabular}{P{0.8cm}|P{1.1cm}|P{2cm}|P{1.5cm}|P{1cm}|P{2.4cm}|P{1.3cm}|P{1cm}}
            \multicolumn{8}{l}{\textbf{Page Title: Joseph Haslet}} \\
            \multicolumn{8}{l}{\textbf{Section Title:} Almanac} \\
            \hline
            \textbf{Year} & \textbf{Office} & \textbf{Subject} & \textbf{Party} &\textbf{\% Votes} & \textbf{Opponent} & \textbf{Party} & \textbf{\% Votes} \\
            \hline
            \cellcolor{yellow}1804 & \cellcolor{yellow}Governor & \cellcolor{yellow}Joseph Haslet& Republican & \cellcolor{yellow}48\% & \cellcolor{yellow}Nathaniel Mitchell & \cellcolor{yellow}Federalist & \cellcolor{yellow}52\%\\
            \hline
            \cellcolor{yellow}1807 & \cellcolor{yellow}Governor & \cellcolor{yellow}{Joseph Haslet} & Republican & \cellcolor{yellow}48\% & \cellcolor{yellow}George Truitt & \cellcolor{yellow}Federalist & \cellcolor{yellow}52\%\\
            \hline
        \end{tabular}
       \label{tab:order_other_a}
    \end{subtable}
    \hfill
         \begin{subtable}[ht]{\textwidth}
        \centering
        \footnotesize
        \captionsetup{justification=raggedright,singlelinecheck=false, font=footnotesize}
        \caption{\textbf{Input Table with Corrected Cells highlighted in Yellow}}
       \vspace{1px}
        \begin{tabular}{P{0.8cm}|P{1.1cm}|P{2cm}|P{1.5cm}|P{1cm}|P{2.4cm}|P{1.3cm}|P{1cm}}
            \multicolumn{8}{l}{\textbf{Page Title: Joseph Haslet}} \\
            \multicolumn{8}{l}{\textbf{Section Title:} Almanac} \\
            \hline
            \textbf{Year} & \textbf{Office} & \cellcolor{cbora}\textbf{Candidate} & \textbf{Party} &\textbf{\% Votes} & \textbf{Opponent} & \textbf{Party} & \textbf{\% Votes} \\
            \hline
            \cellcolor{yellow}1804 & \cellcolor{yellow}Governor & \cellcolor{yellow}Joseph Haslet& \cellcolor{cbora}Republican & \cellcolor{yellow}48\% & \cellcolor{yellow}Nathaniel Mitchell & \cellcolor{yellow}Federalist & \cellcolor{yellow}52\%\\
            \hline
            \cellcolor{yellow}1807 & \cellcolor{yellow}Governor & \cellcolor{yellow}{Joseph Haslet} & \cellcolor{cbora}Republican & \cellcolor{yellow}48\% & \cellcolor{yellow}George Truitt & \cellcolor{yellow}Federalist & \cellcolor{yellow}52\%\\
            \hline
        \end{tabular}
       \label{tab:order_other_b}
    \end{subtable}
     \hfill
    \begin{subtable}[ht]{\textwidth}
        \footnotesize
        \vspace{1px}
        \captionsetup{justification=raggedright,singlelinecheck=false, font=footnotesize}
        \caption{\textbf{Outputs Before and After Input Data Correction}}
        \vspace{2px}
        \begin{tabular}{p{0.99\textwidth}}
            \textbf{T5-base before fix:} Joseph Haslet \markerrword{was elected} Governor in 1804, \markerrword{defeating} Federalist Party candidate \markerrname{Nathaniel Mitchell} in the election of 1807.\\
            \textbf{T5-base after fix:} Joseph Haslet \markerrword{was elected} Governor \markerrword{of} the Republican Party in 1804 and 1807, \markerrword{defeating} the Federalist Party's Nathaniel Mitchell and George Truitt.\\
            \vspace{0.1px}
            \textbf{T5-large before fix: }
            Joseph Haslet \markerrword{was elected} Governor in 1804, \markerrword{defeating} Federalist Party candidate Nathaniel Mitchell in the election of 1807.\\
            \textbf{T5-large after fix:}
            In 1804 and 1807, Republican Joseph Haslet \markerrword{defeated} Federalist Nathaniel Mitchell and Federalist George Truitt. \\
            \vspace{0.1px}
            \textbf{\textsc{Llama 2-13B} before fix:} 1804, Joseph Haslet \markerrword{was elected} governor with 48\% of the vote. In 1807, he \markerrword{was re-elected} with 48\% of the vote.\\

            \textbf{\textsc{Llama 2-13B} after fix:} 1804: Joseph Haslet (Republican) \markerrword{won} with 48\% of the votes. 1807: George Truitt (Federalist) won with 52\% of the votes. \\
        \end{tabular}
        
     \end{subtable}
     \lineacross{}
     \caption{\textbf{ToTTo specific input problems: List of Leader names, as detailed in \Cref{subsubsec: order}}. Correction for this input is including the party name and modifying the header `Subject' to `Candidate'. All models generated Joseph Haslet to be the winning candidate including the \textsc{Llama 2} models that had specific instructions.}
     \label{tab:totto_specific_input_prob_order_three}
     
\end{table*}

\begin{table*}[htbp]
    \lineacross{}
    \captionsetup{justification=raggedright, singlelinecheck=false, font=footnotesize}
    \centering
    \setlength\extrarowheight{2pt}
    \footnotesize
    \begin{subtable}[b]{\linewidth}
        \caption{\textbf{Input Table with Original Cells highlighted in Yellow}}
        \vspace{4px}
         \begin{tabular}{l|l|l|l|l}
        \hline
        \multicolumn{5}{p{0.6\linewidth}}{\textbf{Page Title:} List of members of the United States House of Representatives in the 67th Congress by seniority} \\

        \multicolumn{5}{l}{\textbf{Section Title:} List of Representatives by seniority} \\
        \hline
        \textbf{Rank} & \textbf{Representative} & \textbf{Party} & \textbf{District} & \textbf{Seniority date} \\
        \hline
        \multicolumn{5}{l}{\textbf{Twenty-three non-consecutive terms}} \\
        \hline
        \cellcolor{yellow}1 & Joseph G. Cannon & R & IL-18 & March 4, 1915 \\
        \hline
        \multicolumn{5}{l}{\textbf{Fifteen consecutive terms}} \\
        \hline
        \cellcolor{yellow}2 & Frederick H. Gillett & R & MA-2 & March 4, 1893 \\
        \hline
        \multicolumn{5}{l}{\textbf{Fourteen non-consecutive terms}} \\
        \hline
        \cellcolor{yellow}3 & Henry A. Cooper & R & WI-1 & March 4, 1921 \\
        \hline
        \multicolumn{5}{l}{\textbf{One term}} \\
        \hline
        \cellcolor{yellow}4 & ... & ... & ... & ...\\
        \hline
        \cellcolor{yellow}5 & ... & ... & ... & ...\\
        \hline
        \cellcolor{yellow}433 & ... & ... & ... & ...\\
        \hline
        \cellcolor{yellow}434 & ... & ... & ... & ...\\
        \hline
        \cellcolor{yellow}435 & Adam M. Wyant & PA-22 & WI-1 & March 4, 1921 \\
        \hline
        \multicolumn{5}{l}{\textbf{Members joining the House, after the start of the Congress}} \\
        \hline
        ... & ... & ... & ... & ...\\
        \hline
        \end{tabular}
        \label{tab:long_a}
    \end{subtable}
    
    \begin{subtable}[b]{\linewidth}
        \caption{\textbf{Outputs with Original Input before Input Data Correction}}
        \begin{tabular}{p{0.99\linewidth}}
        \vspace{1px}
        \textbf{T5-large before fix:} The 435th member of the United States House of Representatives in the 65th Congress was a member \markerrcontext{who joined the House after the start of Congress}. \\
        \vspace{1px}

        \textbf{\textsc{Llama2-13B} before fix:} \markerrother{List of Representatives by seniority The table below lists the 100.} \\
       
        \end{tabular}
        
    \end{subtable}
    
    \begin{subtable}[b]{\linewidth}
        \caption{\textbf{Input Table with Corrected Cells highlighted in Yellow}}
        \vspace{4px}
        \begin{tabular}{l|l|l|l|l|l}
        
        \hline
        \multicolumn{6}{p{0.75\linewidth}}{\textbf{Page Title:} List of members of the United States House of Representatives in the 67th Congress by seniority} \\
        \multicolumn{6}{l}{\textbf{Section Title:} List of Representatives by seniority} \\
        \hline
        \textbf{Rank} & \textbf{Representative} & \textbf{Party} & \textbf{District} & \textbf{Seniority date} & \textbf{Term details}\\
        \hline
        \cellcolor{yellow}1 & Joseph G. Cannon & R & IL-18 & March 4, 1915 & {\textbf{Twenty-three non-consecutive terms}}\\
        \hline
        \cellcolor{yellow}2 & Frederick H. Gillett & R & MA-2 & March 4, 1893 & {\textbf{Fifteen consecutive terms}} \\
        \hline
        \cellcolor{yellow}3 & Henry A. Cooper & R & WI-1 & March 4, 1921 &  {\textbf{Fourteen non-consecutive terms}}\\

        \hline
        \cellcolor{yellow}4 & ... & ... & ... & ...&  {\textbf{One term}}\\
        \hline
        \cellcolor{yellow}5 & ... & ... & ... & ... & {\textbf{One term}}\\
        \hline
        \cellcolor{yellow}433 & ... & ... & ... & ... &  {\textbf{One term}}\\
        \hline
        \cellcolor{yellow}434 & ... & ... & ... & ... &  {\textbf{One term}}\\
        \hline
        \cellcolor{yellow}435 & Adam M. Wyant & PA-22 & WI-1 & March 4, 1921 &  {\textbf{One term}} \\
        \hline

        \end{tabular}
        \label{tab:long_fix}
    \end{subtable}
    
    \begin{subtable}[b]{\linewidth}
        \caption{\textbf{Outputs After Input Data Correction}}
        \begin{tabular}{p{0.99\linewidth}}
        \vspace{1px}
            \textbf{T5-large after fix:} There are 435 members of the United States House of Representatives in the 65th Congress. \\
            \vspace{1px}
            \textbf{\textsc{Llama2-13B} after fix:} 
            The table shows that there are 435 representatives in the 65th Congress, and they are ranked according to their seniority. \\
        \end{tabular}
    \end{subtable}
    \lineacross{}
    
    \caption{\textbf{Longer Table input problems, as detailed in \Cref{subsubsec: longer}}. \newline
    (a). Original tabular input had 435 rows with term details and other details from nested headers (irrelevant details). \newline
    (b). The fine-tuned T5-large model hallucinated the output text with the irrelevant header information (nested header issue). \textsc{Llama 2-13B} also struggled to produce the right text and generated incomplete output. \textsc{Llama 2-7B} produced longer garbage output. \newline
    (c). First, we fixed the nested header by creating a separate column. Then, we only passed 20 records (first 10 rank and 426 to 435 rank) along with title information. We did not pass the irrelevant term details to the corrected data. \newline
    (d). This simplified input records with relevant header details (Rank) fixed the errors in both models.}
    \label{tab:long_table}
    
\end{table*}

\end{appendices}

\end{document}